\RequirePackage{amsmath}

\documentclass[twocolumn]{svjour3}          
\smartqed  
\usepackage{graphicx}
\usepackage{times}
\usepackage{epsfig}
\usepackage{epstopdf}
\usepackage[labelfont=bf]{caption}
\usepackage{tabularx}
\usepackage[table]{xcolor}
\usepackage{fixltx2e}
\usepackage{epstopdf}
\AppendGraphicsExtensions{.tif}

\newcommand{\etal}{\textit{et al.}}
\newcommand{\tabincell}[2]{\begin{tabular}{@{}#1@{}}#2\end{tabular}}
\newcolumntype{H}{>{\setbox0=\hbox\bgroup}c<{\egroup}@{}}

\newcommand{\spacebelowtab}{\vspace{-5mm}}
\newcommand{\spaceabovetab}{\vspace{-3mm}}

\newcommand{\spaceabovesubsection}{\vspace{-5mm}}

\usepackage{makecell}
\usepackage[colorlinks,linkcolor=red,urlcolor=blue,anchorcolor=blue,citecolor=green]{hyperref}
\usepackage{amsfonts}
\usepackage{color}
\newcolumntype{H}{>{\setbox0=\hbox\bgroup}c<{\egroup}@{}}
\usepackage{colortbl}
\definecolor{editcolor}{RGB}{0,0,0}
\definecolor{babypink}{rgb}{0.96, 0.76, 0.76}
\definecolor{mygray}{RGB}{221,221,221} 
\usepackage{bm}
\usepackage{graphicx}
\usepackage{array}
\usepackage{booktabs} 
\usepackage{bm}
\usepackage{arydshln}
\usepackage{colortbl}
\usepackage{xcolor}
\usepackage{enumitem}
\usepackage{multirow} 
\usepackage[caption=false]{subfig}
\usepackage[sort,nocompress]{cite}
\usepackage{bbding}
\journalname{International Journal of Computer Vision}
\begin{document}

\title{EAN: Event Adaptive Network for Enhanced Action Recognition
}

\author{Yuan~Tian,
	Yichao~Yan \Envelope,
	Guangtao~Zhai \Envelope,
	Guodong~Guo,
	and Zhiyong~Gao
}

\institute{Y. Tian, G. Zhai, and Z. Gao are with the Institute of Image Communication and Network Engineering, Shanghai Jiao Tong University, Shanghai, China. E-mail: \{ee\_tianyuan,zhaiguangtao,zhiyong.gao\}@sjtu.edu.cn.
Y. Yan is with the AI Institute, Shanghai Jiao Tong University, Shanghai, China. E-mail: yanyichao@sjtu.edu.cn.
G. Guo is with Baidu. E-mail: guoguodong01@baidu.com.
\Envelope~denotes the corresponding author.
}
\vspace{-5mm}
\date{\textsf{Received: 9 September 2021 / Accepted: 20 July 2022}}

\maketitle

\begin{abstract}
  Efficiently modeling spatial-temporal information in videos is crucial for action recognition. To achieve this goal, state-of-the-art methods typically employ the convolution operator and the dense interaction modules such as non-local blocks. However, these methods cannot accurately fit the diverse events in videos. On the one hand, the adopted convolutions are with fixed scales, thus struggling with events of various scales. On the other hand, the dense interaction modeling paradigm only achieves sub-optimal performance as action-irrelevant parts bring additional noises for the final prediction. In this paper, we propose a unified action recognition framework to investigate the dynamic nature of video content by introducing the following designs. First, when extracting local cues, we generate the spatial-temporal kernels of dynamic-scale to adaptively fit the diverse events. Second, to accurately aggregate these cues into a global video representation, we propose to mine the interactions only among a few selected foreground objects by a Transformer, which yields a sparse paradigm. We call the proposed framework as \textit{Event Adaptive Network} (EAN) because both key designs are adaptive to the input video content. To exploit the short-term motions within local segments, we propose a novel and efficient \textit{Latent Motion Code} (LMC) module, further improving the performance of the framework. Extensive experiments on several large-scale video datasets, \textit{e.g.}, Something-to-Something V1\&V2, Kinetics, and Diving48, verify that our models achieve state-of-the-art or competitive performances at low FLOPs. \textit{Codes are available at: \url{https://github.com/tianyuan168326/EAN-Pytorch}.} 

\end{abstract}

 \begin{figure}[t]
	
	\subfloat[ \textit{Pushing something so that it falls off the table}]{%
		\centering
		\includegraphics[width=8cm]{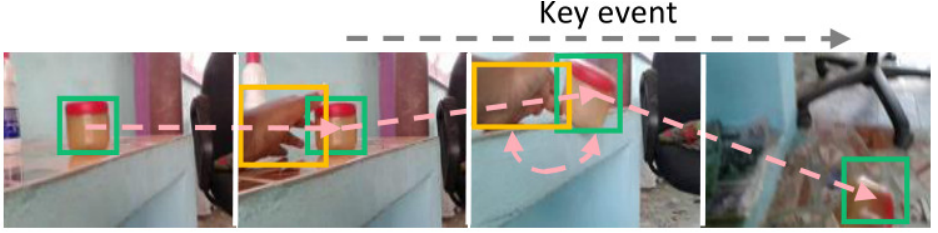}
	}
	\vspace{-3mm}
	\hfill
	
	\subfloat[\textit{Moving something towards the camera}]{%
		\centering
		\includegraphics[width=8cm]{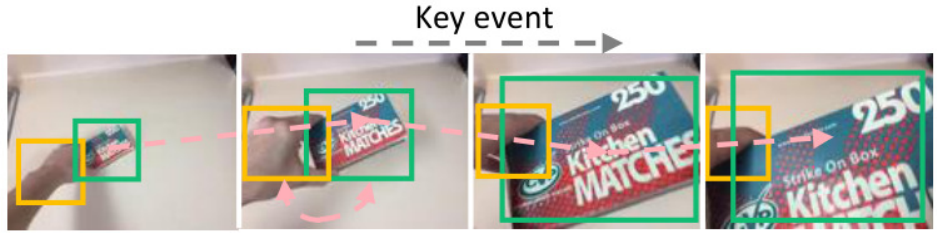}  
	}
	\caption {Two examples from the Something-Something dataset~\cite{goyal2017something}.
		The objects (\textit{i.e.}, can, box, and hand) and events have diverse spatial-temporal scales in different videos.
		Therefore, convolution kernels with adaptive scales can better fit them.
		Moreover, the interactions among them are naturally sparse,
		which can be accurately and efficiently modeled by a dedicated sparse model.
		{The objects, object interactions, and key events are indicated by the colored boxes, \textcolor{babypink}{pink} dotted arrows, and \textcolor{gray}{gray} dotted arrows, respectively.}
	}
	\label{fig_1}
\end{figure}

\section{Introduction}
\label{intro}

  Video action recognition is an open challenge in computer vision, drawing increasing attention in both research and industrial communities, because of its fundamental role for tremendous applications, \textit{e.g.}, human behavior monitoring~\cite{cherian2019second}\cite{chen2021sportscap}\cite{tian2019video}, video surveillance~\cite{ferryman2000visual}, anomaly events analysis~\cite{bensch2017spatiotemporal}\cite{lu2019fast}, to name a few.
  It goes beyond the recognition performed on single images and depends on comprehensively modeling both (1) the local spatial-temporal cues and (2) the global object interactions in videos.
  
  Many previous methods~\cite{wang2018temporal}\cite{lin2019tsm}\cite{zhou2018temporal}\cite{tran2015learning}\cite{carreira2017quo}~\cite{tran2018closer}~\cite{feichtenhofer2019slowfast} achieve promising performance by only modeling the local spatial-temporal cues.
  However, these networks are typically built with convolutions, whose scales are usually empirically determined and kept fixed for different input videos.
  Indeed, by designing multi-scale networks, such as in ResNet
~\cite{he2016deep}, Inception networks~\cite{szegedy2016rethinking}, and Res2Net~\cite{gao2019res2net}, the models are equipped with convolution kernels of diverse scales.
  But, these architectures are still static, not adapting to the various events within videos.
  We illustrate this challenge in Fig.~\ref{fig_1}.
  There naturally arises a question - \textit{can we design a dynamic architecture that adaptively fits the events in each video?}

  Additionally,
  recognizing the actions in videos needs to reason about the interactions among the objects.
  Although the local interactions can be well captured by the convolutions, there are always some non-local interactions that can only be observed from a global view. For example, in Fig.~\ref{fig_1} (a), the key interaction is ``{the can is moved towards the ground across several frames''.}
  Modeling interactions like this requires global reasoning capability, which is beyond the function of convolution.
  To model the global information, dense interaction models~\cite{wang2018non}\cite{bertasius2021space}\cite{fan2021multiscale} calculate the paired correlations at all positions, which inevitably introduce the background noise signals. In contrast, the sparse models~\cite{wang2018videos}\cite{materzynska2020something} are more accurate because they only target the action-relevant regions.
  Nevertheless, the inefficiency and the error accumulation caused by their embedded object detector are nontrivial to resolve.
  Moreover, the utilized heavy detector hinders the end-to-end training of the whole system.
  Therefore, there arises another question - \textit{can we model the global object interactions sparsely without relying on a heavy object detector?
  }

  In this paper, we answer both questions with \textit{yes}, by carefully designing several spatial-temporal modeling modules.
  \textit{First}, we propose an {E}vent {A}daptive  Block (EAB) to enhance the convolution operators with scale-adaptive modeling capability. Particularly, this block perceives the scale information of the key events within the input video, and then dynamically synthesizes the spatial-temporal kernel. Since the scale of the kernel is not fixed, it is unfeasible to represent it with a single trainable tensor. Instead, we reformulate it as a soft fusion of several fixed-scale spatial- or temporal-convolution kernels.
  Since the synthesized kernel is customized to the input video, the local event cues within the video are better modeled.
  Moreover, the prevalent architectures, \textit{e.g.}, R(2+1)D CNNs~\cite{tran2018closer} and Inception-Nets~\cite{szegedy2016rethinking} can be viewed as a special case of the proposed EAB.
  \textit{Second}, we propose a {S}parse {O}bject {I}nteraction {Tr}ansformer (SOI-Tr) to build sparse interaction graphs by adaptively selecting the most important objects involved in the actions. Concretely, given the deep video features, an embedded object localization network first outputs several saliency maps, each of which corresponds to an object. Then, a shallow Transformer~\cite{vaswani2017attention} is used to model the long-range interactions among this small number of objects. Thanks to the feature-level detection scheme, this module gets rid of the heavy detector and is end-to-end trainable, which is more effective and efficient than the previous models~\cite{wang2018videos}\cite{materzynska2020something}.
  
  In addition to the two spatial-temporal modeling modules above, we further propose a novel Latent Motion Code (LMC) module to efficiently exploit the short-term motion information within local video segments.
  Specifically, the low-level motion cues within each segment, \textit{i.e.}, RGB differences, are first encoded into a compact latent space. Then, the high-order motion information is reasoned in this space.
  The motion information further facilitates the discriminating capability of our method for some hard action cases.
  
  We incorporate the proposed three modules into a unified ConvNet called Event Adaptive Network (EAN).
  By following a series of efficient network designs, the proposed EAN is highly efficient.
  The whole framework can be jointly optimized following the sparse sampling strategy proposed in TSN~\cite{wang2018temporal}.
  We emphasize our contributions as follows:
  \begin{itemize}
  	
  	\item A novel Event Adaptive Block (EAB) is proposed to generate the video-adaptive spatial-temporal convolution kernel of dynamic scale, demonstrating superior local spatial-temporal modeling capability.
  	Moreover, our approach is the very first work to generate dynamic spatial-temporal convolution kernels for video data.
  	
  	\item A Sparse Object Interaction Transformer (SOI-Tr) is developed to accurately reason the global interactions among the sparse foreground objects, without relying on bounding box annotations or external object detectors.
  	
  	\item A novel and efficient Latent Motion Code (LMC) module is devised to capture the short-term motion information within local video segments in a latent space.
  	
  	\item By incorporating the proposed EAB, SOI-Tr, and LMC into the off-the-shelf 2D CNNs, \textit{i.e.}, 2D ResNet, we build up a strong yet efficient video action recognition framework called Event Adaptive Network (EAN). Our models achieve state-of-the-art or competitive results on several large-scale video datasets, \textit{i.e.}, Something-Something V1\&V2~\cite{goyal2017something}, Kinetics~\cite{carreira2017quo}, and Diving48~\cite{li2018resound}.
  \end{itemize}

\section{Related work} 
\label{sec:related}

\begin{figure*}[!tbp]
	\centerline{\includegraphics[width=15cm]{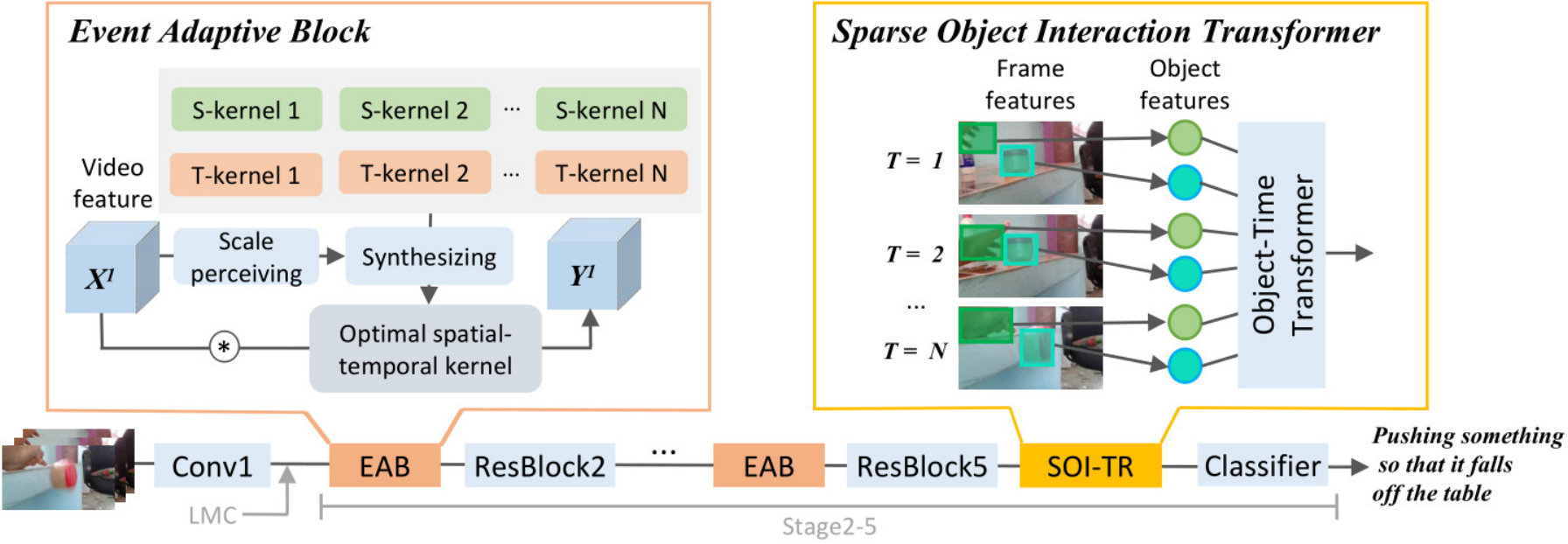}}
	\caption{
		\textbf{Event Adaptive Network}.
		Our framework aims to simultaneously model local spatial-temporal information and global object interactions by incorporating two novel modules, \textit{i.e.}, {E}vent {A}daptive {B}lock (EAB) and {S}parse {O}bject {I}nteraction {Tr}ansformer (SOI-Tr), into the 2D ResNet backbone CNN. The EAB first perceives the scale of local events and then dynamically synthesizes video-adaptive spatial-temporal kernels from spatial (S) or temporal (T) kernels of fixed scales.
		The SOI-Tr specializes in global interactions among sparse foreground objects by leveraging a Transformer.
		Besides, a latent motion code (LMC) module is adopted to efficiently exploit short-term motion information within local segments.
		Our proposed framework is an end-to-end hybrid model that uses both convolution and self-attention.
	}
\vspace{-4mm}
	\label{fig_framework}
\end{figure*}

\noindent \textbf{Deep Action Recognition.}
Two-stream CNNs~\cite{simonyan2014two}\cite{feichtenhofer2016convolutional}\cite{feichtenhofer2020deep} are the earliest works on deep action recognition.
Later, many methods~\cite{wang2018temporal}\cite{zhou2018temporal}\cite{lin2019tsm}\cite{liu2020tam}\cite{liu2020teinet}\cite{luo2019grouped}\cite{li2020tea}\cite{wang2020tdn}\cite{wu2021coarse}\cite{khowaja2020semantic}\cite{tian2021self} enhance the 2D CNNs with various temporal modules and achieve promising results.
To simultaneously learn the temporal dynamics along with the spatial representations in videos, 3D networks, \textit{e.g.}, C3D network~\cite{tran2015learning}, I3D~\cite{carreira2017quo}, 3D-ResNet~\cite{hara2017learning}\cite{tian2020self}, R(2+1)D CNNs~\cite{tran2018closer}, and Slowfast networks~\cite{feichtenhofer2019slowfast}, have also recently gained much attention.
Our framework is built upon the 2D CNNs due to their better efficiency.

\noindent \textbf{Multi-scale CNNs.}
Many modern image CNN architectures, \textit{e.g.}, Inception networks~\cite{szegedy2015going}\cite{szegedy2016rethinking}\cite{szegedy2017inception}, Res2Net~\cite{gao2019res2net}, incorporate the multi-scale design for obtaining better representations.
For the video tasks, Zhang \etal~\cite{zhang2020pan} proposed the various-timescale inference pooling to observe videos across various timescales. 
TEA~\cite{li2020tea} extends the Res2Net block with temporal modeling capability. However, all these architectures are static, while our method is dynamic and adaptive to the input video.

\noindent \textbf{Dynamic Convolution.}
Jia~\etal~\cite{jia2016dynamic} first proposed the concept of dynamic filter.
Latter, several works~\cite{yang2019condconv}\cite{chen2020dynamic} in image tasks attempt to dynamically generate aggregation
weights and use them to combine a set of convolutional kernels. More recently, TANet~\cite{liu2020tam} generalizes this idea to temporal modeling for the video recognition task. However, the generated temporal kernel is shared across all channels, demonstrating limited modeling capability and performance.
In contrast, our method generates the full \textit{spatial-temporal} kernel, whose parameters are specified for each channel.

\noindent \textbf{Object Interaction models.}
Ma~\etal~\cite{ma2018attend} employed the LSTM to build the object-object interaction graph.
Wang~\etal~\cite{wang2018videos} utilized the Graph Convolutional Networks (GCNs) to perform object relationship reasoning.
~Materzynska~\etal~\cite{materzynska2020something} proposed a sparse semantically grounded subject-object graph representation.
All these methods rely on an object detector or external bounding box annotations to determine the regions of the objects.
Without leveraging any explicit object region information, Non-local Neural Networks~\cite{wang2018non} try to model every pairwise interaction \textit{densely} in the feature space.

\noindent \textbf{Vision Transformer.}
Recently, many works~\cite{dosovitskiy2020image}\cite{touvron2020training}\cite{bertasius2021space}\cite{fan2021multiscale}\cite{girdhar2021anticipative}\cite{arnab2021vivit}\cite{zhang2021vidtr}\cite{bulat2021space}\cite{cong2021spatial} apply the Transformer architecture~\cite{vaswani2017attention}
to image/video tasks by unfolding the visual signal or its feature map to a sequence of tokens.
Although their global modeling capability is inherently superior to the convolution-based methods, these models are computationally-expensive due to the dense self-attention mechanism. Similar to us, Girdhar \etal~\cite{girdhar2019video} and Plizzari~\etal~\cite{plizzari2021skeleton} also leverage a lightweight Transformer architecture to model the interactions among the selected key regions of the input video. However, they either rely on an object region proposal network (RPN) to produce dense proposal regions or an external computationally-heavy keypoint extractor to locate the human keypoints. In contrast, our object representation is sparse and is produced by a lightweight three-layer CNN.

\noindent \textbf{Short-term Motion Representation.}
Previous works of two-stream action recognition frameworks~\cite{simonyan2014two}\cite{feichtenhofer2016convolutional} use optical flow maps as a complement to RGB inputs.
Both conventional or CNN-based optical flow estimation methods~\cite{zach2007duality}\cite{ilg2017flownet} \cite{ranjan2017optical}\cite{sun2018pwc} can be adopted in this framework.
Recent works~\cite{zhang2020pan} \cite{wang2020tdn} propose several lightweight modules to produce task-specific short-term motion representations, which can be jointly optimized with the action recognition network. For example, PAN~\cite{zhang2020pan} proposes to use the difference of the low-level features between the adjacent frames as a novel motion cue named Persistence of Appearance (PA).
More recently, TDN~ \cite{wang2020tdn} uses a short-term module to map the RGB difference motion signals into compact features and fuse the features with that produced by the backbone network.
In contrast, our proposed latent motion code (LMC) module exploits high-order motion information in a latent space, which is more effective and also efficient.

\section{Approach}
\label{approach}
In this work, we propose a novel video action recognition framework called Event Adaptive Network (EAN), as shown in Fig.~\ref{fig_framework}.
The network is built by inserting several Event Adaptive Blocks (EABs) and a {S}parse {O}bject {I}nteraction {Tr}ansformer (SOI-Tr) into different stages of the 2D ResNet backbone CNN. Moreover, a Latent Motion Code (LMC) module is adopted to exploit the short-term motion information within local video segments.
All components in our framework are differentiable and the proposed EAN is end-to-end trainable.

\spaceabovesubsection
\subsection{Event Adaptive Block}\label{EAB}

\begin{figure}[!t]
	\centering
	\includegraphics[width=8cm]{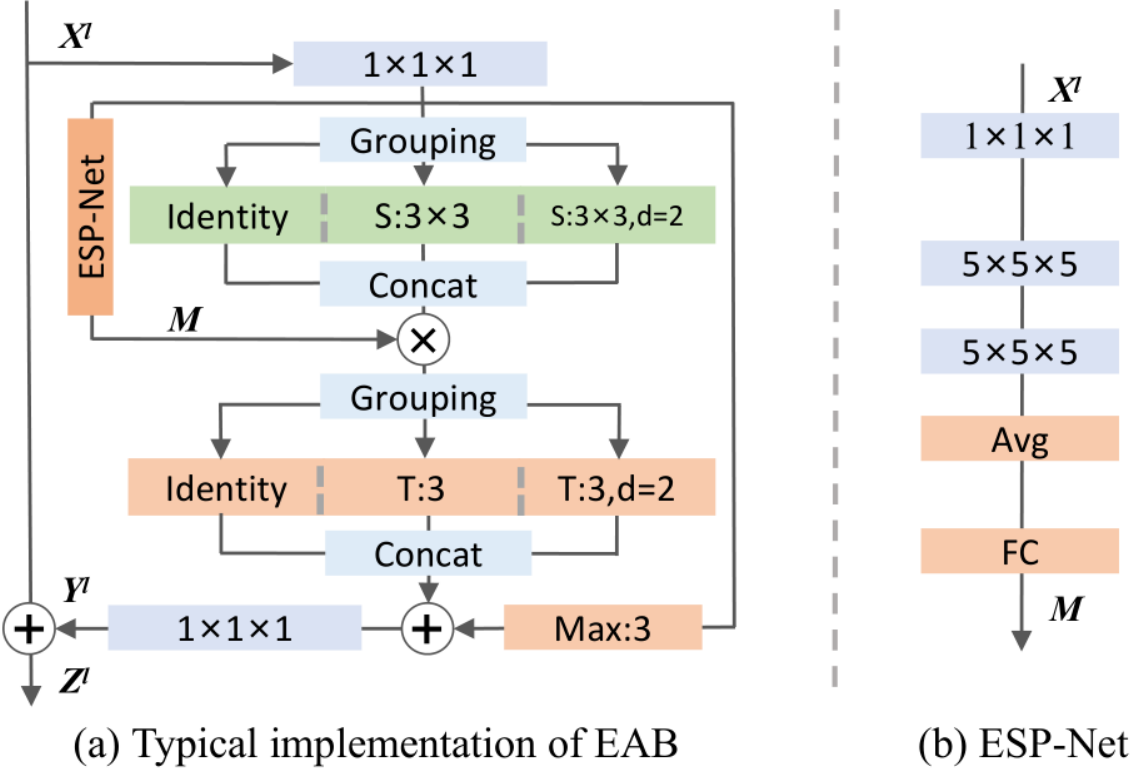}
	\caption {(a) \textbf{EAB} of maximum receptive filed size 5$\times$5$\times$5. (b) A zoom-in of ESP-Net.
		“S:3$\times$3, d=2” represents a 2D spatial convolution with kernel size 3 and dilation size 2. “T” indicates a 1D temporal convolution. “Max:3” denotes the 3D max-pooling operator with kernel size 3. ``Avg'', $\oplus$ and $\otimes$ denote the average pooling, the element-wise addition, and the broadcasting channel-only matrix multiplication, respectively. ``FC'' denotes a fully connected layer.
	}
	\vspace{-4mm}
	\label{fig_ean}
\end{figure}

Event Adaptive Block (EAB) aims to generate the spatial-temporal kernel to adaptively model the local cues within the input video, as shown in Fig.~\ref{fig_ean}.
We start from an approximated formulation of the optimal kernel for the video, and then implement the formulation as an efficient block.

\textbf{An approximated formulation for the optimal kernel.}
Formally, given an input video feature $X^l$ with channel number $C$, which is the output of the $l$-th ($l \in [1,4]$) stage of the backbone CNN. We first assume that there exists an optimal spatial-temporal kernel $\hat{F}$ that accurately fits the key elements (\textit{i.e.}, objects and events) of the video. This kernel transforms the input $X^l$ into an output tensor $Y^l$ of the same shape by convolution:
\begin{align} 
Y^l = X^l * \hat{F}.
\label{eq1}
\end{align}
Both the scale and the parameters of the $\hat{F}$ can adapt to videos with different contents.
Because the accurate shape of $\hat{F}$ is unknown, we cannot easily implement it as a trainable fixed-scale convolution kernel. Instead, we propose to solve the surrogate problem, \textit{i.e.}, approximating the produced $Y^l$.
We achieve this by leveraging a group of fixed-scale spatial or temporal convolutions:
\begin{align}\label{mk_multi}
	&Y^l={\cup}_{i=1}^G \{M \otimes {\cup}_{j=1}^G X^l_j * F_s^{(2j-1)}\}_i * F_t^{(2i-1)},
\end{align}
where $F_s^{(2j-1)}$ and $F_t^{(2i-1)}$ represent the spatial convolution with kernel size $(2j-1) \times (2j-1)$ and the temporal convolution with kernel size $2i-1$, respectively.
Each convolution is performed on a group of features for reducing the computation cost.
$G$ denotes the group number, $\cup$ denotes the channel concatenating operation, and $X^l_j = X^l[j \cdot c : (j+1) \cdot c]$, where $c = C/G$.
$\otimes$ denotes the channel-wise broadcasting matrix multiplication operation.
$M$ denotes the fusion matrix that relates the spatial and temporal convolutions, and is estimated by the Event Scale Perceiving Network (ESP-Net):
\begin{align}
	M = \textsf{ESP-Net} (X^l), M \in \mathbb{R}^{C \times C}.
\end{align}

As formulated in Eq.~(\ref{mk_multi}), $M$ dynamically gates the spatial information flowed into each temporal convolution. By choosing different $M$, we can mimic the previous hand-crafted video architectures.
For example, by only activating the matrix elements connecting the spatial and temporal kernels with the same size, the proposed formulation degenerates to the (2+1)D convolutions. Moreover, the multi-scale spatial-only or temporal-only convolutions are also the special cases of it.

\textbf{Event Scale Perceiving Network (ESP-Net).}
It is well known that the scale information is embodied in the spatial-temporal context, which encodes the rich semantics \textit{w.r.t} the shapes of objects and the dynamics of events.
Thus,
ESP-Net is implemented as a lightweight 3D network with a small channel number but a large receptive field,
as shown in Fig.~\ref{fig_ean} (b). Specifically, a 1$\times$1$\times$1 3D convolution layer is first adopted to reduce feature channels of the input tensor $X^l$ by 16 times.
Then, the video context features are extracted with two 3D convolutions with kernel size $5 \times 5 \times 5$ and stride size $2 \times 2 \times 2$. Subsequently, the average pooling operation is utilized to only reserve the channel dimension of the tensor,
and globally aggregate the event scale information of the input video.
Finally, a linear transformation layer followed by a reshaping operation is utilized to produce $M$.

\textbf{Implementation of EAB.}
We wrap the above procedure into an Event Adaptive Block (EAB).
This block is defined as:
$
Z^l = Y^l + X^l,
$
where ${Y^l}$ is given in Eq.~(\ref{mk_multi}) and “$+~X^l$” denotes a residual connection~\cite{he2016deep}.
The residual connection allows us to insert the proposed block into any pre-trained model such as ResNet, without breaking its initial behavior (\textit{e.g.}, when the weights of the last Conv layer in EAB are initialized as zeros).
An example of EAB with maximum receptive field size $5\times5\times5$ is illustrated in Fig.~\ref{fig_ean} (a).
Bottleneck design is introduced for reducing the computation complexity, \textit{i.e.}, we first reduce the feature channel number by four times through 1$\times$1$\times$1 convolutions.
The spatial convolutions are followed by batch normalization (BN) and ReLU non-linearity.
We also introduce a max-pooling branch as a complement for convolution.
To further reduce the parameter and the computational complexity, we replace the convolution of large kernel size with dilated convolution.

\textbf{Further discussion with dynamic convolution.}
Dynamic Convolution~\cite{chen2020dynamic} proposes to decouple the dynamic convolution as the attentions over several static convolutions.
Nevertheless, our method is dedicated to video data while they are only for image data.
In addition to that, there are several other significant differences between our method and them.
\textit{First}, our method is not merely \textit{context-adaptive} but also \textit{scale-adaptive}. More concretely, during the kernel generation procedure, dynamic convolution uses a global average pooling (GAP) operation to extract the global context information as the first step. In contrast, we reserve the additional spatial-temporal dimensions and utilize the 3D convolutions to extract the scale information of the objects and events.
\textit{Second}, the convolutions adopted in our method are with various kernel sizes to adapt to the events of various scales, while that in dynamic convolution are with the same kernel size.
\textit{Third}, the element of $M$ in our method is specified for each channel, while the attention weight of dynamic convolution is shared across all channels of the convolution.

\spaceabovesubsection
\subsection{Sparse Object Interaction Transformer}\label{sec_method_interaction}

\begin{figure}[!thb]
	\centering
	\includegraphics[width=8cm]{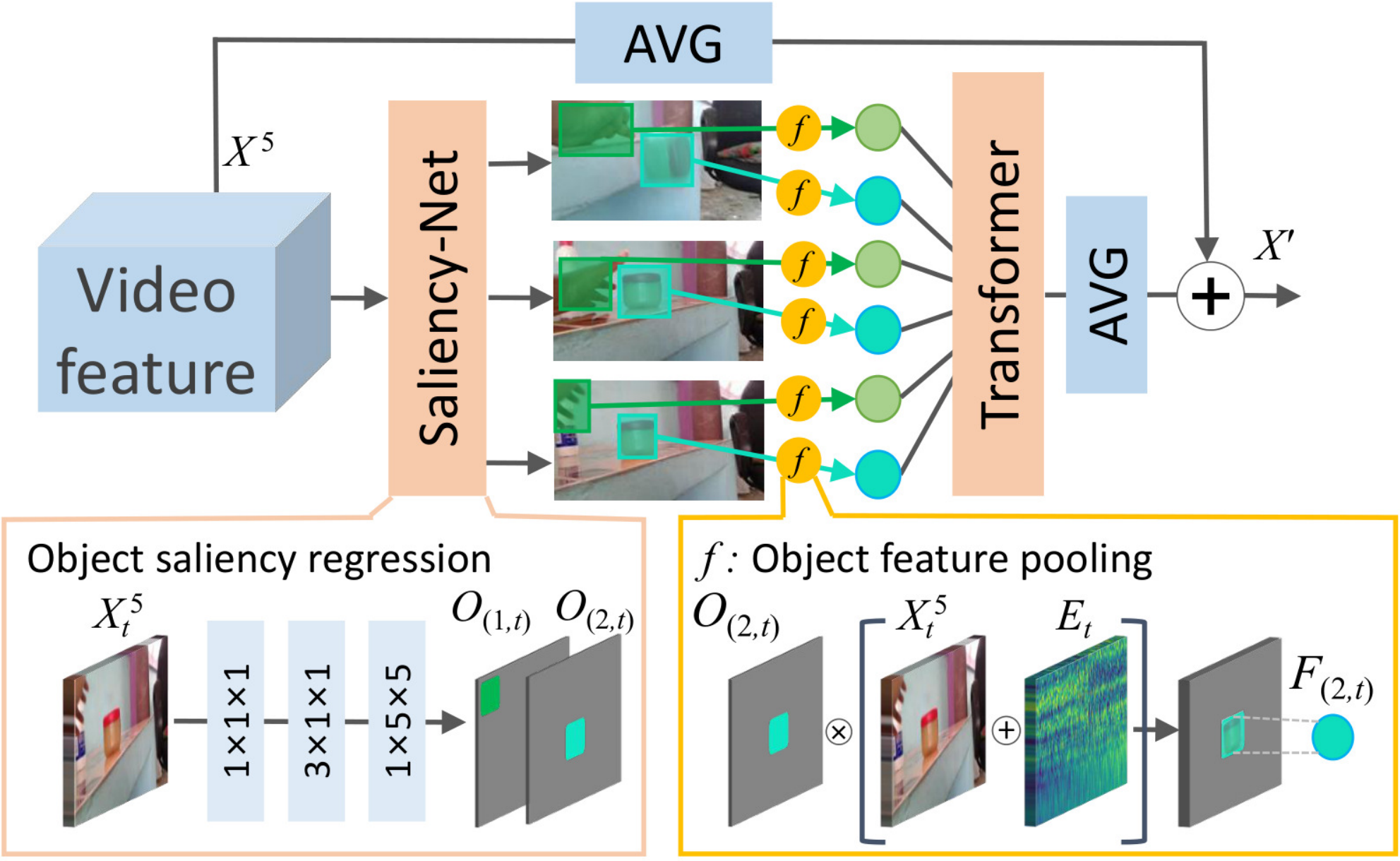}
	\caption {
		The architecture of the proposed Sparse Object Interaction Transformer (\textbf{SOI-Tr}).
		$O_{(2,t)}$ represents the saliency map for the second object in the $t$-{th} frame.
		$E_t$ denotes the positional embedding.
		We illustrate the feature map as the original RGB frame for better intuitive understanding.
	}
		\vspace{-4mm}
	\label{fig_soitr}
\end{figure}

The proposed EAB only captures the local information of the video, lacking the global modeling capability. Therefore, we propose a Sparse Object Interaction Transformer ({SOI-Tr}) to aggregate the local action cues into a global representation, as shown in Fig.~\ref{fig_soitr}. To make the modeling procedure more accurate for the specific input video, we only mine the interactions among the foreground objects in each frame, which are localized on the fly in the feature space.

Given the output feature of the $5$-{th} stage of the backbone CNN $X^5 \in \mathbb{R}^{ C \times T \times W \times H}$, where $C$ denotes the channel number, $T$ denotes the temporal length, $W \times H$ represent the spatial scales, we model SOI-Tr as follows:

\textit{(1) Localizing the foreground objects.}
The location of each object is represented as a two-dimensional saliency map, whose spatial scale is equivalent to that of $X^5$, \textit{i.e.}, $W \times H$.
We use a small fully convolutional network (FCN) termed Saliency-Net to regress the object saliency maps in parallel:
\begin{align}
	{O} = \textsf{Saliency-Net}(X^5), {O} \in \mathbb{R}^{ N \times T \times W \times H},
\end{align}
where $N$ denotes the maximum number of the foreground objects in one frame.
We empirically set $N = 4$ as most actions only involve less than four objects.

\begin{figure*}[!tb]
	
	\centerline{\includegraphics[width=16cm]{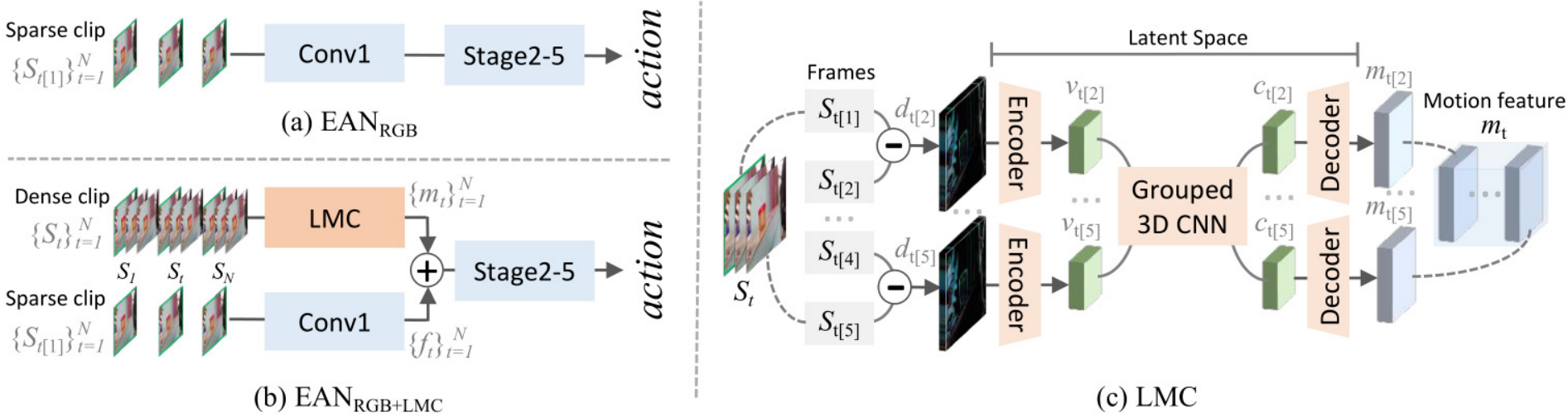}}
	\caption {(a) EAN$_\textnormal{RGB}$ model only takes the sparsely sampled clip as the input. 
		(b) EAN$_\textnormal{RGB+LMC}$ model takes both the densely sampled clip and the sparse clip as the input. (c) Zooming-in of the Latent Motion Code (LMC) module, which transforms a local video segment $S_t$ to a compact motion feature $m_t$.}
	\label{fig_ean_motion}
		\vspace{-4mm}
\end{figure*}

\textit{(2) Pooling the object features.}
We first denote the saliency map for the $n$-th object in the $t$-th frame as ${O}_{(n,t)} \in \mathcal{R}^{ W \times H}$.
Then, the feature representation of the object is produced by spatially weighting the input video feature with the saliency map:
\begin{align}
	{F}_{(n,t)} = \textsf{SSUM}((E_t + X^5_t) \odot {O}_{(n,t)}), {F}_{(n,t)} \in \mathbb{R}^{C},
\end{align}
where $\odot$ denotes the broadcasting element-wise multiplication, $\textsf{SSUM}$ denotes the summation across spatial dimensions.
$E_t$ denotes a learnable spatially positional embedding with the same shape as $X^5_t \in \mathbb{R}^{C \times W \times H} $.

\textit{(3) Modeling the object interactions.}
With the object-level features, we model the global interactions among them using a Transformer:
\begin{align}
	{F'} =  \textsf{Transformer}({F}), {F'} \in \mathbb{R}^{N \times T  \times C }.
\end{align}
The produced ${F'}$ is with the same shape as $F$.

\textit{(4) Enhancing the global video representation.}
Finally, we perform average-pooling on the original video features and the interaction features, yielding the global video representation ${X}'$.
It should be mentioned that the SOI-Tr module also adopts the bottleneck designing with a channel compressing factor of four.

\textbf{Saliency-Net.}
This network is implemented as a light-weight four-layer CNN followed by a spatial Softmax layer, where the first layer reduces the the input channel number by a factor of eight. The second layer is a 3D convolution with kernel size 3$\times$1$\times$1, which detects the moving objects with obvious motions. The third layer is a 2D convolution with a larger spatial kernel size 5$\times$5, which localizes objects more accurately by considering the context information.

\textbf{Transformer.}
The transformer architecture used in our framework is built by stacking two residual blocks, where each block includes a multi-head ${QKV}$ self-attention module. Different from the vanilla Transformer, we mainly remove the classification token and replace the Layer Normalization with Batch Normalization, following the practices proposed in~\cite{srinivas2021bottleneck}.

\spaceabovesubsection
\subsection{Latent Motion Code Module}
The proposed EAB and SOI-Tr modules can already extract the action cues from the input video clip effectively. We insert several EABs and a SOI-Tr into the 2D ResNet backbone, building the EAN$_\textnormal{RGB}$ model, as shown in Fig.~\ref{fig_ean_motion} (a).
EAN$_\textnormal{RGB}$ already recognizes the actions from the sparsely sampled video clip effectively. Nevertheless, some subtle action cues are inevitably lost during the sampling procedure. To alleviate this issue, we sample the video clip more densely, and introduce a novel Latent Motion Code (LMC) module to efficiently mine the motion cues within the local segments of this dense clip, as shown in Fig.~\ref{fig_ean_motion} (c). When equipping EAN$_\textnormal{RGB}$ with the LMC module, we build an improved EAN$_\textnormal{RGB+LMC}$ model, as shown in Fig.~\ref{fig_ean_motion} (b).
Although the EAN$_\textnormal{RGB+LMC}$ takes more frames as input, it is also with high efficiency, due to the adopted latent motion modeling scheme and early feature fusion strategy.

\textbf{Frame Sampling Strategy}.
Following the previous works
\noindent~\cite{wang2016temporal}\cite{zhang2020pan}\cite{wang2020tdn}, we uniformly divide the original long video into several groups and then select a segment from each group. Specifically, the input video is first divided into $N$ groups with equal length. $N$ is 8 or 16 for different computational budgets. During the training procedure, five adjacent frames are randomly chosen from each group as a 5-frame segment. The $N$ segments form a dense clip
$ { \{ {S_t} \}_{t=1}^N  } $. The first frames of each segment form a sparse clip $ { \{ {S_{t[1]}} \} _{t=1}^N  }  $.

\textbf{Latent Motion Code Module (LMC)}.
This module aims to transform the short-term motion information within each local segment into a single compact motion feature, as shown in Fig.~\ref{fig_ean_motion} (c).
Given an input segment $ S_t $, we model the motion information within it as follows:

\textit{(1) Calculating RGB difference maps.}
We obtain the low-level motion cue, \textit{i.e.}, RGB difference map, by subtracting every two consecutive frames:
\begin{align}
	d_{t[i]} =  S_{t[i]} - S_{t[i-1]}, d_{t[i]} \in \mathbb{R}^{3 \times 224 \times 224}, i \in [2,5].
\end{align}

\textit{(2) Encoding motion from RGB to latent space.}
Due to the high redundancy between the consecutive frames, the produced difference map is naturally sparse and contains many near-zero values.
To simultaneously improve the compactness of the signal and also filter out the task-unrelated motion information, we use a learnable encoder to transform it into a high-dimensional latent space.
Specifically, we divide $d_{t[i]}$ into $7\times7 = 49$ patches,
where each patch is of shape $3 \times 32 \times 32$.
Then, we compress these three-dimensional patches into 128-element latent vectors, and the vectors form a latent map of size $7\times7$:
\begin{align}
	v_{t[i]} = \textsf{H}_\textsf{e} (d_{t[i]}), v_{t[i]} \in \mathbb{R}^{128 \times 7 \times 7},
\end{align}
where the encoder $\textsf{H}_\textsf{e}$ is implemented as a linear layer that is shared by all patches. The input and output dimensions of the layer are $3 \times 32 \times 32 = 3072$ and 128, respectively.

\textit{(3) Modeling high-order motion in latent space.}
The latent map is with low resolution and thus can be efficiently processed by 3D convolutions:
\begin{align}
	\{ c_{t[i]} \}_{t=2}^5 = \textsf{H}_\textsf{m} (\{ v_{t[i]} \}_{t=2}^5), c_{t[i]} \in \mathbb{R}^{128 \times 7 \times 7},
\end{align}
where $\textsf{H}_\textsf{m}$ is implemented as two stacking 3D convolutions with kernel size 3 and group size 16.
We call the produced $c_{t[i]}$ as latent motion code (LMC) because it captures the high-order motion information in the latent space.

\textit{(4) Decoding motion from latent to feature space.}
Through another linear transformation, LMCs can be decoded into the feature space.
Following TDN~\cite{wang2020tdn}, we align the dimensions of the decoded features with the features from the Conv1 stage.
Concretely, we decode the vector in each spatial position of LMC into a feature patch of size $16 \times 8 \times 8$, and these $7 \times 7$ patches form a motion feature map of shape $16 \times 56 \times 56$:
\begin{align}
	m_{t[i ]} = \textsf{H}_\textsf{d} ( c_{t[i]} ), m_{t[i ]} \in \mathbb{R}^{16 \times 56 \times 56},
\end{align}
where the decoder $\textsf{H}_\textsf{d}$ is implemented as a linear layer with the input dimension of 128 and the output dimension of $16 \times 8 \times 8$, respectively.
Finally, the motion feature for segment $S_t$ is constructed by stacking the motion feature maps along the channel dimension:
\begin{align}
	m_{t} = [m_{t[2]} ; m_{t[3]} ; m_{t[4]} ; m_{t[5]}], m_{t} \in \mathbb{R}^{64 \times 56 \times 56}.
\end{align}

\textbf{EAN$_\textnormal{RGB+LMC}$ architecture}.
As shown in Fig.~\ref{fig_ean_motion} (b), for each segment $S_t$, we add the motion feature $m_{t}$ produced by the LMC module to the Conv1 feature of the first frame $S_{t[1]}$:
\begin{align}
	f_{t} = m_{t} + \textsf{Conv1} (S_{t[1]}).
\end{align}
Then, the fused features of each segment are fed to the remained stages of EAN for predicting the action category score:
\begin{align}
	action =  \textsf{Stage2-5} (\{  f_{t}  \}^N_1),
\end{align}
where \textsf{Conv1} and \textsf{Stage2-5} are indicated in Fig.~\ref{fig_framework}.

\section{Experiments}

\noindent \textbf{Datasets.}
We evaluate our method on several large-scale video datasets with different properties, requiring our models to understand different aspects of action recognition task.

{\textit{Something-Something}} includes V1~\cite{goyal2017something} and V2~\cite{mahdisoltani2018fine} versions, which are two large-scale crowd-sourcing video datasets for action recognition.
There are about 110k (V1) and 220k (V2) videos covering 174 fine-grained action categories with diverse objects and scenes,
focusing on humans performing pre-defined basic actions.
In this dataset, the actions are performed with different objects so that models are required to understand the basic actions instead of recognizing the appearance of the objects or the background scenes.
Moreover, the spatial and the temporal scales of the objects and the events vary hugely across different videos, as shown in Fig.~\ref{fig_1},
which is suitable for verifying the flexible spatial-temporal modeling ability of the proposed method.

{\textit{Kinetics}}~\cite{carreira2017quo} is a challenging human action recognition dataset, which contains 400 and 600 human action classes.
This dataset includes human-object interactions such as playing instruments, as well as human-human interactions such as shaking hands and hugging.
Compared to the temporal reasoning required by the actions in Something-Something, the actions in this dataset heavily rely on the appearance of the objects.
We evaluate our models on the trimmed version to evaluate its capacity in modeling the appearances and the interaction among objects.
The experiments are conducted on the validation set of Kinetics-400~\cite{carreira2017quo} because there are many well-known baseline methods.

{\textit{Diving48}}~\cite{li2018resound} includes more than 18K video clips for 48 unambiguous diving classes.
This proves to be a challenging task for modern action recognition systems as dives include three stages (takeoff, flight, entry) and thus require modeling of long-term temporal dynamics.
This requires both multi-scale temporal modeling and the perceiving of long-range dependencies.
Therefore, we conduct experiments on this dataset to verify the multi-scale spatial-temporal modeling ability of our method.
We report the accuracy on the first version of the official validation split, which has been adopted by several previous methods.\footnote{http://www.svcl.ucsd.edu/projects/resound/Diving48\_\{train/test\}.json}

\noindent \textbf{Implementation Detail}
We implement our model in Pytorch, and we adopt ResNet50~\cite{he2016deep} pretrained on ImageNet~\cite{deng2009imagenet} as the backbone.
Following previous works~\cite{kwon2020motionsqueeze}\cite{li2020tea}, we also insert temporal convolutions with kernel size 3 and the motion excitation (ME) module proposed in \cite{li2020tea} before each 3$\times$3 convolutions of bottleneck layers of the original ResNet50, aiming to enhance its basic temporal modeling ability. We also incorporate these changes into all baselines in the ablation study for a fair comparison. 
The parameters within the EABs and SOI-Tr are randomly initialized.
For the spatial dimension of the sampled clips, the short-side of the frames are resized to $256$ and then cropped to $224\times224$. We perform random cropping and flipping as data augmentation during training.
It's worth mentioning that we do not perform horizontal flipping on the moving direction related action classes such as \textit{``moving something from left to right''}.
We train the network with a batch size of 64 and optimize it using SGD with an initial learning rate of 0.01 for 40 epochs, and decay it by a factor of 10 for every 10 epochs. The total training epochs are about 70.
The dropout ratio is set to 0.5. The weight decay is set to $5e^{-4}$ and $1e^{-4}$ for Something/Diving48 and Kinetics-400, respectively.\footnote{We adopt the same hyper-parameter settings as \href{https://github.com/MCG-NJU/TDN}{the official codebase of TDN}
for a fair comparison.}

\spaceabovesubsection
\subsection{Comparison with State-of-the-Arts}

 \textbf{Something V1 and V2.}
We first compare our method with the other state-of-the-art approaches on Something V1 and Something V2 datasets, as shown in Tab.~\ref{tab_something_sota}.
The previous approaches are divided into four groups: 3D CNNs, object interaction modeling enhanced 3D CNNs, 2D CNNs, and 2D CNNs enhanced with short-term motion representation.

\begin{table*}[!ht]
	\centering
	\caption{Comparison to state-of-the-arts on Something-Something V1\&V2 datasets.
	Following TDN~\cite{wang2020tdn}, we adopt the \textit{1-clip and center-crop} inference scheme where only a center crop of 224$\times$224 from a single clip is used for evaluation.
	\textsubscript{8F} and \textsubscript{16F} indicate the sampling segment number of the input video is 8 and 16, respectively.
	The result of EAN$_{\operatorname{En(RGB)}}$ is produced by averaging the predicted action scores from the EAN$_{\operatorname{8F(RGB)}}$ and EAN$_{\operatorname{16F(RGB)}}$ models, which follows TSM~\cite{lin2019tsm}.
	$-$ indicates the paper didn't provide the results.
	}
	\scalebox{1.0}{
		\renewcommand{\arraystretch}{1.0}
		\setlength{\tabcolsep}{0.8mm}
		
		\begin{tabular}{ccccccccccc}
			\Xhline{2\arrayrulewidth}
			\multirow{2}{*}{\tabincell{c}{\textbf{Method}} } &
			\multirow{2}{*}{\tabincell{c}{\textbf{Backbone}} } &
			\multirow{2}{*}{\tabincell{c}{\textbf{Pre-train}} } &
			\multirow{2}{*}{\tabincell{c}{\textbf{Frames}} } &
			\multirow{2}{*}{\tabincell{c}{\textbf{GFLOPs}} } &
			\multicolumn{2}{c}{\textbf{Something V1}}  & \multicolumn{2}{c}{\textbf{Something V2}} \\
			&&&&& \textbf{Top1 (\%)} & \textbf{Top5 (\%)}  & \textbf{Top1 (\%)} & \textbf{Top5 (\%)}\\
			\hline 
			\multicolumn{2}{l}{\tabincell{l}{\textbf{3D CNNs:}}} \\
			
			I3D~\cite{carreira2017quo}  & 3D-ResNet50 & Kinetics & 32$\times$2 & 306 & 41.6 &72.2 &-&- \\ 
			Non-local I3D~\cite{wang2018non}  & 3D-ResNet50 & Kinetics & 32$\times$2 & 336 & 44.4&76.0 &-&- \\ 
			\arrayrulecolor{mygray}\cdashline{1-9}[5pt/3pt]
			ECO(En)~\cite{zolfaghari2018eco} & BNInc+3D-ResNet18 & Kinetics & 92  & 267  & 46.4 &- &-&- \\  
			\arrayrulecolor{gray}\cdashline{1-9}[5pt/3pt]
			S3D-G~\cite{xie2018rethinking}  & InceptionV1 & ImageNet & 64  & 71  & 48.2&78.7 &-&- \\
			\hline
			\multicolumn{2}{l}{\tabincell{l}{\textbf{3D CNNs + Object interaction:}}} \\
			
			GCN + Non-local~\cite{wang2018videos} & 3D-ResNet50 & Kinetics & 32$\times$2 & 606 & 46.1&76.8 &-&- \\
			I3D + STIN + OIE~\cite{materzynska2020something} & I3D & Kinetics & 32 & 154 & - &- & 60.2 &84.4 \\
			\hline 	
			\multicolumn{2}{l}{\tabincell{l}{\textbf{2D CNNs:}}} \\
			TSN~\cite{wang2016temporal}  & BN-Inception & ImageNet & 8  & 16  & 19.5 &- &33.4 & -\\ 
			MultiScale TRN~\cite{zhou2018temporal}  & BN-Inception & ImageNet & 8  & 16  & 34.4 &63.2 &48.8 &77.6  \\ 
			\arrayrulecolor{gray}\cdashline{1-9}[5pt/3pt]
			TSM$_{\operatorname{8F}}$~\cite{lin2019tsm}   & ResNet-50 & Kinetics & 8  & 33  & 45.6 &74.2 &58.8&85.4  \\ 
			TSM$_{\operatorname{16F}}$~\cite{lin2019tsm}   & ResNet-50 & Kinetics & 16  & 65  & 47.2 &77.1 &63.4& 88.5 \\ 
			\arrayrulecolor{gray}\cdashline{1-9}[5pt/3pt]
			TANet$_{\operatorname{8F}}$~\cite{liu2020tam}  & ResNet-50 & ImageNet & 8  & 33 & 46.5 &75.8& 60.5&86.2 \\
			TANet$_{\operatorname{16F}}$~\cite{liu2020tam}  & ResNet-50 & ImageNet & 16  & 66 & 47.6& 77.7& 62.5&87.6 \\
			TANet$_{\operatorname{En}}$~\cite{liu2020tam}  & ResNet-50 & ImageNet & 8+16  & 99 & 50.6& 79.3 & -&- \\
			\arrayrulecolor{gray}\cdashline{1-9}[5pt/3pt]
			TEINet$_{\operatorname{8F}}$~\cite{liu2020teinet}  & ResNet-50 & ImageNet & 8  & 33 & 47.4 &-& 61.3&- \\
			TEINet$_{\operatorname{16F}}$~\cite{liu2020teinet}  & ResNet-50 & ImageNet & 16  & 66 & 49.9& -& 62.1&- \\
			TEINet$_{\operatorname{En}}$~\cite{liu2020teinet}  & ResNet-50 & ImageNet & 8+16  & 99 & 52.5& -& 65.5&89.8 \\
			\arrayrulecolor{gray}\cdashline{1-9}[5pt/3pt]
			STM~\cite{jiang2019stm}   & ResNet-50 & ImageNet & 8$\times$30 & 990 & 49.2&79.3 &62.3&88.8 \\ 
			STM~\cite{jiang2019stm}   & ResNet-50 & ImageNet & 16$\times$30 & 2010 & 50.7&80.4 &64.2&89.8 \\ 
			\arrayrulecolor{gray}\cdashline{1-9}[5pt/3pt]
			GST$_{\operatorname{8F}}$~\cite{luo2019grouped}  & ResNet-50 & ImageNet & 8  & 29  & 47.0&76.1 &-&- \\
			GST$_{\operatorname{16F}}$~\cite{luo2019grouped}  & ResNet-50 & ImageNet & 16  & 59  & 48.6&77.9 & 62.6&87.9\\ 
			\arrayrulecolor{gray}\cdashline{1-9}[5pt/3pt]
			TEA$_{\operatorname{8F}}$~\cite{li2020tea}  & ResNet-50 & ImageNet & 8  & 35  & 48.9&78.1 &-&-\\
			TEA$_{\operatorname{16F}}$~\cite{li2020tea}  & ResNet-50 & ImageNet & 16  & 70  & 51.9&80.3 &-&-\\ 
			TEA~\cite{li2020tea}  & ResNet-50 & ImageNet & 16$\times$30 & 2100 & 52.3&81.9 &65.1&89.9\\  
			\arrayrulecolor{gray}\cdashline{1-9}[5pt/3pt]
			\arrayrulecolor{gray}\cdashline{1-9}[5pt/3pt]

			EAN$_{\operatorname{8F(RGB)}}$(Ours) & ResNet-50 & ImageNet & 8 & 36  &{51.9} &{79.5} &{63.5}&{88.2}\\ 
			EAN$_{\operatorname{16F(RGB)}}$(Ours) & ResNet-50 & ImageNet & 16  & 72  & {53.4} & {81.4} &{64.6}&{89.1} \\ 
			EAN$_{\operatorname{En(RGB)}}$(Ours)  & ResNet-50 & ImageNet & 8+16 & 108  & \textbf{55.8} & \textbf{83.1} &\textbf{66.6}&\textbf{89.9}\\
			
			\hline 	
			\multicolumn{2}{l}{\tabincell{l}{\textbf{2D CNNs + Short-term motion:}}} \\
			TRN$_{\operatorname{RGB+Flow}}$~\cite{zhou2018temporal}  & BN-Inception & ImageNet & 8$\times$7  & - & 42.0 &- &55.5 & 83.1  \\ 
			TSM$_{\operatorname{RGB+Flow}}$~\cite{lin2019tsm}  & ResNet-50 & ImageNet & 16$\times$7  & - & 52.6& 81.9& 66.0& 90.5&  \\
			\arrayrulecolor{gray}\cdashline{1-9}[5pt/3pt]
			PAN$_{\operatorname{8F(RGB+PAN)}}$~\cite{zhang2020pan} & ResNet-50 & ImageNet &  8$\times$5  		& 68 & 50.5 &79.2 &63.8 &88.6 \\
			PAN~\cite{zhang2020pan} & ResNet-101 & ImageNet &  (8$\times$5)$\times$2  &  503 &55.3 &82.8 &66.5 &90.6 \\
			\arrayrulecolor{gray}\cdashline{1-9}[5pt/3pt]
			TDN$_{\operatorname{8F(RGB+SDM)}}$~\cite{wang2020tdn} & ResNet-50 & ImageNet &  8$\times$5  		& 36 & 52.3 & 80.6 &64.0 &88.8 \\
			TDN$_{\operatorname{16F(RGB+SDM)}}$~\cite{wang2020tdn} & ResNet-50 & ImageNet &  16$\times$5  		& 72 & 53.9& 82.1 &65.3 &89.5 \\
			TDN$_{\operatorname{En(RGB+SDM)}}$~\cite{wang2020tdn} & ResNet-50 & ImageNet &  (8+16)$\times$5  		& 108 & 55.1 & 82.9 &67.0& 90.3 \\
			\arrayrulecolor{gray}\cdashline{1-9}[5pt/3pt]
			EAN$_{\operatorname{8F(RGB+LMC)}}$(Ours) & ResNet-50 & ImageNet &  8$\times$5  		& 37 & {53.4} & {81.1} &{65.2} &{89.4} \\
			EAN$_{\operatorname{16F(RGB+LMC)}}$(Ours) & ResNet-50 & ImageNet &  16$\times$5  		& 74 & {54.7}& {82.3} &{66.6} &{90.3} \\
			EAN$_{\operatorname{En(RGB+LMC)}}$(Ours) & ResNet-50 & ImageNet &  (8+16)$\times$5  		& 111 & \textbf{57.2} & \textbf{83.9} &\textbf{68.8}& \textbf{91.4} \\
			\Xhline{2\arrayrulewidth}
		\end{tabular}
	}
	
	\label{tab_something_sota}
\end{table*}

Our method outperforms all methods built with 3D convolutions and meanwhile achieves higher efficiency. For example, compared with Non-local I3D~\cite{wang2018non}, our EAN\textsubscript{8F(RGB+LMC)} model achieves 8.8\% higher Top1 accuracy (44.4\% \textit{vs.} 53.2\% on Something V1) with only $\sim 11\%$ computational cost.

\begin{table*}[!th]
	
	\caption{Comparison to state-of-the-arts on Kinetics-400 dataset.
	Following TDN~\cite{wang2020tdn}, we adopt the \textit{10-clip and 3-crop} inference scheme where three crops of 256$\times$256 frames and 10 clips are used for testing.
	Therefore, the computational cost of the same model here is 30$\times$ heavier than that on Something datasets.
	$-$ indicates the paper didn't provide the results.}
	\centering
	\scalebox{1.02}{
		\renewcommand{\arraystretch}{1.0}
		\setlength{\tabcolsep}{2.0mm}
		\begin{tabular}{ccccccc}
			\Xhline{2\arrayrulewidth}
			\textbf{Method} & \textbf{Backbone} & \textbf{Pre-train} & \textbf{Frames} & \textbf{GFLOPs} & \textbf{Top1 (\%)} & \textbf{Top5 (\%)} \\
			\hline
			ARTNet~\cite{wang2018appearance}  & ResNet-18 & ImageNet & 16$\times$250 & 23.5$\times$250 & 70.7 &89.3 \\ 
			I3D~\cite{carreira2017quo}  & Inception V1 & ImageNet & 64$\times$N/A & 108$\times$N/A & 72.1 &90.3 \\ 
			I3D~\cite{carreira2017quo}  & Inception V1 & None & 64$\times$N/A & 108$\times$N/A & 67.5 &87.2 \\ 
			I3D+NL~\cite{wang2018non}  & 3D-ResNet-101 & ImageNet & 32$\times$60 & 359$\times$60 & 77.7&93.3 \\ 
			
			ECO(En)~\cite{zolfaghari2018eco} & BNInc\&3D-ResNet-18 & None & 92  & 267  & 70.0 &- \\ 
			
			SlowOnly~\cite{feichtenhofer2019slowfast} & 3D-ResNet-50 & None & 8$\times$30 & 41.9$\times$30 & 74.8 & 91.6 \\
			SlowFast~\cite{feichtenhofer2019slowfast} & 3D-ResNet-50 & None & (4+32)$\times$30 & 36.1$\times$30 & 75.6 & 92.1 \\ 
			SlowFast+NL~\cite{feichtenhofer2019slowfast} & 3D-ResNet-101 & None & (16+64)$\times$30 & 234$\times$30 & \textbf{79.8} & \textbf{93.9} \\ 
			
			\hline 	
			TSN~\cite{wang2016temporal}  & BN-Inception & ImageNet & 25$\times$10 & 53$\times$10 & 69.1 &88.7 \\ 
			TSN~\cite{wang2016temporal}  & Inception v3 & ImageNet & 25$\times$10 & 80$\times$10 & 72.5 &90.2 \\ 
			
			R(2+1)D~\cite{tran2018closer}  & ResNet-34 & None & 32$\times$10 & 152$\times$10 & 72.0 & 90.0 \\ 
			
			TSM~\cite{lin2019tsm}   & ResNet-50 & ImageNet & 8$\times$30 & 33$\times$30 & 74.1 &- \\ 
			TSM~\cite{lin2019tsm}   & ResNet-50 & ImageNet & 16$\times$30 & 65$\times$30 & 74.7 &-  \\ 
			
			STM~\cite{jiang2019stm}   & ResNet-50 & ImageNet & 16$\times$30 & 67$\times$30 & 73.7&91.6  \\ 
			
			TEINet~\cite{liu2020teinet}  & ResNet-50 & ImageNet & 8$\times$30 & 33$\times$30 & 74.9&91.8 \\ 
			TEINet~\cite{liu2020teinet}  & ResNet-50 & ImageNet & 16$\times$30 & 66$\times$30 & 76.2&92.5 \\
			
			TANet~\cite{liu2020teinet}  & ResNet-50 & ImageNet & 8$\times$30 & 43$\times$30 & 76.1&92.3 \\ 
			TANet~\cite{liu2020teinet}  & ResNet-50 & ImageNet & 16$\times$12 & 86$\times$12 & 76.9&92.9 \\
			TEA~\cite{li2020tea}  & ResNet-50 & ImageNet & 16$\times$30 & 70$\times$30 & 76.1&92.5 \\ 
			PAN~\cite{zhang2020pan}  & ResNet-50 & ImageNet &  (8$\times$5)$\times$2 & 270 & 75.3&92.4 \\
			TDN$_{\operatorname{8F(RGB+SDM)}}$~\cite{wang2020tdn}  & ResNet-50 & ImageNet & (8$\times$5)$\times$30 & 36$\times$30 & 76.6&92.8 \\
			TDN$_{\operatorname{16F(RGB+SDM)}}$~\cite{wang2020tdn}  & ResNet-50 & ImageNet & (16$\times$5)$\times$30 & 72$\times$30 & 77.5&93.2 \\
			TDN$_{\operatorname{En(RGB+SDM)}}$~\cite{wang2020tdn}  & ResNet-50 & ImageNet & (8+16)$\times$5$\times$30 & 108$\times$30 & 78.4&93.6 \\
			TDN$_{\operatorname{En(RGB+SDM)}}$~\cite{wang2020tdn}  & ResNet-101 & ImageNet & (8+16)$\times$5$\times$30 & 198$\times$30 & \textbf{79.4}&\textbf{94.4} \\
			\hline 
			EAN$_{\operatorname{8F(RGB+LMC)}}$(Ours) 	& ResNet-50 & ImageNet & (8$\times$5)$\times$30 &  37$\times$30 & 77.1 &93.3 \\
			EAN$_{\operatorname{16F(RGB+LMC)}}$(Ours)& ResNet-50 & ImageNet & (16$\times$5)$\times$30 &  74$\times$30 & 78.3 &93.7 \\
			EAN$_{\operatorname{En(RGB+LMC)}}$(Ours)& ResNet-50 & ImageNet & (8+16)$\times$5$\times$30 &  111$\times$30 & \textbf{79.0} &\textbf{94.1} \\
			\Xhline{2\arrayrulewidth}
		\end{tabular}
	}
	
	\label{tab_k400_sota}
\end{table*}

We also compare our method with the two methods~\cite{materzynska2020something}\cite{wang2018videos} that first detect the objects of the input frames in the RGB space and then model the object interactions.
Although we do not use the pretrained object detector or extra object bounding box annotations to get the proposal regions, our method still significantly outperforms them.
Specifically, our improvements over GCN + Non-local~\cite{wang2018videos} and I3D + STIN + OIE~\cite{materzynska2020something} are \textbf{11.1\%} (on Something V1) and \textbf{8.6\%} (on Something V2), respectively, in terms of the Top1 recognition accuracy.
This proves the superiority of the end-to-end object detection scheme and the Transformer architecture adopted in our method.

As for the 2D CNN-based methods, we compare our EAN$_{\operatorname{RGB}}$ architecture with them for a fair comparison, where only one frame is sampled from each segment.
Our models achieve the best performance under all settings of different input frame numbers.
The performances of TSN and TRN are relatively inferior to other methods because both the two methods only model the temporal information upon the highest-level feature maps from the backbone CNN. TEA is superior to all other 2D CNNs because it explores multi-scale spatial-temporal information. Compared with TEA, our method outperforms it consistently with the different input frame numbers. When using 8 and 16 input frames, the improvements are 3.0\% and 1.5\% on Something V1 dataset. The reason is that the multi-scale architecture of TEA is based on the hand-crafted Res2Net, which is static and not adaptive to the video. In contrast, the spatial-temporal modeling architecture of our method is dynamic and adaptive.

We further compare the improved EAN$_\textnormal{RGB+LMC}$ architecture with the other recent 2D CNNs that also take advantage of the short-term motion information, where 5 adjacent frames are sampled from each segment. Compared with the optical flow-based methods, \textit{i.e.}, TRN$_{\operatorname{RGB+Flow}}$ and TSM$_{\operatorname{RGB+Flow}}$, 
our smallest model EAN$_{\operatorname{8F(RGB+LMC)}}$ already outperforms them by 11.2\% and 0.6\%, respectively. It's worth noting that the computational complexity of our LMC motion feature produced from the input video of 40 frames is only 1.1 GFLOPs, while the computational complexity of FlowNet2.0~\cite{ilg2017flownet} is 2006 GFLOPs for the same video.
In other words, the proposed LMC module is 1823$\times$ more efficient than optical flow modality, while achieving better performance for the action recognition task.

By averaging the predictions from EAN$_{\operatorname{8F(RGB+LMC)}}$ and EAN$_{\operatorname{16F(RGB+LMC)}}$, the resulted model EAN$_{\operatorname{En(RGB+LMC)}}$ boosts the action recognition performance to a new state-of-the-art level, \textit{i.e.}, 57.2\% (\textbf{$\bigtriangleup$ +2.1\%}) on Something V1 and 68.8\% (\textbf{$\bigtriangleup$ +1.8\%}) on Something V2, when using the recent method TDN as the anchor.
Compared with the PAN model, the improvement of our method is 1.9\% on Something V1, even though that PAN adopts a much heavier backbone network, \textit{i.e.}, 2D-ResNet101.

		\begin{table}[!tb]
			\caption{Comparison of different models in terms of input clip setting and their key modules, \textit{i.e.}, local spatial-temporal modeling module, global aggregation module and short-term motion modeling module.
				PA, TSM and ME indicate the appearance of persistence module~\cite{zhang2020pan}, the temporal shift module~\cite{lin2019tsm} and 
				the motion excitation module~\cite{li2020tea}, respectively. 
				L/SDM represents the long/short-term temporal difference modules in TDN~\cite{wang2020tdn}.
				AVG indicates the spatial-temporal average pooling operation.
				The model performances on Something V1 are also reported.
			}
		\centering
			\renewcommand{\arraystretch}{1.0}
			\setlength{\tabcolsep}{0.7mm}
			\begin{tabular}{cccccc}
				\Xhline{2\arrayrulewidth}
				\multirow{2}{*}{\textbf{Model}}& \multicolumn{1}{c}{\textbf{Input clip}} & \multicolumn{3}{c}{\textbf{Key components}} &
				\multirow{2}{*}{\textbf{Top1 (\%)}}
				 \\
				& segment$\times$frame  & Local   & Global  & Motion\\  
				\hline
				TSM\textsubscript{8F} & 8 $\times$ 1 & TSM & AVG &- & 45.6 \\
				ResNet baseline & 8 $\times$ 1 & ME & AVG &- & 48.6 \\
				
				EAN\textsubscript{8F(RGB)} & 8 $\times$ 1 & EAB & SOI-Tr &- &51.9 \\ 
				\hline
				PAN\textsubscript{8F(RGB+PAN)} & 8 $\times$ 5 & TSM & AVG & PA & 50.5 \\ 
				TDN\textsubscript{8F(RGB+SDM)} & 8 $\times$ 5 & LDM & AVG & SDM & 52.3\\
				EAN\textsubscript{8F(RGB+LMC)} & 8 $\times$ 5 & EAB & SOI-Tr & LMC & \textbf{53.4} \\ 
				
				\Xhline{2\arrayrulewidth}
			\end{tabular}
		
		\label{tab_method_compare}
	
\end{table}

Furthermore, we summarize the key differences among our adopted ResNet baseline, our variant models and
other recent relevant methods, as shown in Tab.~\ref{tab_method_compare}.
Particularly, TDN also uses a short-term temporal difference module (SDM) to exploit short-term motion information in the low-level feature space, and fuse the motion features into the backbone in the early stage. Nevertheless, our method outperforms TDN by 1.1\% on Something V1.
The consistent improvements of our method over the other methods strongly justify the superiority of the proposed event scale adaptive spatial-temporal modeling paradigm by EAB, sparse object interaction modeling scheme by SOI-Tr, and high-order motion representation in latent space by LMC.

 \textbf {Kinetics-400.}
To verify that our method also effectively captures rich object appearance cues and the interactions among them, we compare our method with other state-of-the-art results on the Kinetics-400, as shown in Tab.~\ref{tab_k400_sota}.
When compared with the methods based on 2D CNNs, our method outperforms all of them when using the same backbone network, and demonstrates a better trade-off between the action recognition accuracy and the computational complexity.
For example, when equipped with the same ResNet-50 backbone, our method outperforms the recent method TDN by 0.6\%.
When adopting the ResNet-101 backbone, TDN shows the strongest result among all 2D CNNs. Nevertheless, this also increases the computation cost of TDN, which is even close to the 3D CNN method, \textit{i.e.}, SlowFast + NL network.
Our EAN models achieve the best complexity performance trade-off among all state-of-the-art methods.

\begin{table}[!b]
\caption{
	Comparison to state-of-the-arts on Diving48.
	We adopt the \textit{single clip or twice clips} inference schemes where a center crop of 224$\times$224 from a single clip or twice clips is used.
	$-$ indicates the paper didn't provide the results.
}
	\centering
	\scalebox{1.0}{
		\renewcommand{\arraystretch}{1.0}
		\setlength{\tabcolsep}{3.0mm}
		\begin{tabular}{cccc}
			\Xhline{2\arrayrulewidth}
			\textbf{Method} & \textbf{Pre-train} & \textbf{Frames} & \textbf{Top1 (\%)} \\  
			\hline
			TSN (from \cite{li2018resound}) & ImageNet & 8 & 16.7 \\ 
			TRN (from \cite{li2018resound})& ImageNet & 8 & 22.8 \\ 
			C3D (from \cite{li2018resound})& ImageNet & 64 & 27.6 \\ 
			R(2+1)D (from \cite{bertasius2018learning}) & Kinetics&- & 28.9 \\ 
			P3D (from~\cite{luo2019grouped}) & ImageNet&16 & 32.4 \\ 
			C3D (from~\cite{luo2019grouped}) & ImageNet &16& 34.5 \\ 
			Kanojia~\etal~\cite{kanojia2019attentive} & ImageNet &64& 35.6 \\ 
			TEA-ResNet50~\cite{li2018resound} & ImageNet &16&  36.0 \\
			CorrNet-101~\cite{wang2020video} & - &32$\times$10& 38.6 \\ 
			GST ~\cite{luo2019grouped} & ImageNet &16& 38.8  \\  
			\hline  
			Ours & ImageNet &16&  {40.4} \\ 
			Ours & ImageNet &16$\times$2&  \textbf{41.7} \\ 
			\Xhline{2\arrayrulewidth}
		\end{tabular}
	}
	
	\label{tab_diving48_sota}
\end{table}

 \textbf {Diving48.}
To prove that our method can model subtle fine-grained motion cues, we test our method on Diving48.
This dataset requires modeling the subtle body motions in long-short terms and includes much fewer videos compared with Something-Something and Kinetics.
We input 16 frames to the network and sample {two} clips from the video during inference.
The results are shown in Tab.~\ref{tab_diving48_sota}.
Our method outperforms the recent state-of-the-art GST~\cite{luo2019grouped} when using single clip (\textbf{$\bigtriangleup$ +1.6\%}) or twice clips (\textbf{$\bigtriangleup$ +2.9\%}) as the input videos.

\spaceabovesubsection
\subsection{Ablation Studies for EAN}
We conduct extensive ablation studies on Something V1~\cite{goyal2017something} dataset to demonstrate the superiority of the proposed framework by answering the following questions.
The variant models in this section are derived from the EAN$_{\operatorname{8F(RGB)}}$ model.
The input clip is always with 8 frames.

\textit{\textbf{Q1: Are the proposed EAB and SOI-Tr effective and necessary?}}
As mentioned in Sec.~\ref{approach}, in our framework, the EAB extracts more accurate local spatial-temporal representation and 
the SOI-Tr derives global object interaction representation from the video. To confirm that both two representations are effective and necessary for a high-performance action recognition framework, we conduct ablation experiments. Specifically, we equip ResNet baseline with the two proposed modules separately and analyze their impact on the performance.

\begin{table}[!htbp]
\caption{
	Comparison of the performance of using different spatial-temporal modeling modules.
}
	\centering
	\setlength{\tabcolsep}{1.5mm}
	\renewcommand{\arraystretch}{1}
	
	\begin{tabular}{lcccc}
		\Xhline{2\arrayrulewidth}
		\multirow{2}{*}{\tabincell{c}{\textbf{Method} } } &\multirow{2}{*}{\tabincell{c}{\textbf{Param}} }  &\multirow{2}{*}{\tabincell{c}{\textbf{FLOPs}} } & \multicolumn{2}{c}{\textbf{Something V1}}  \\
		&&& \scriptsize{\textbf{Top1 (\%)}} & \scriptsize{\textbf{Top5 (\%)}} \\ 
		\hline
		ResNet baseline	  &24.0M &33.1G & 48.6 & 77.5  \\
		ResNet+EABs &29.5M &35.3G   & 50.8 & 78.4 \\
		ResNet+SOI-Tr &30.3M &33.8G   & 49.3 & 77.9 \\
		\hline
		\textbf{ResNet+EABs+SOI-Tr} &36.0M &36.1G & \textbf{51.9} & \textbf{79.5} \\
		\Xhline{2\arrayrulewidth}
		
	\end{tabular}
	\label{tab_ab_pathway}
\end{table}

As shown in Tab.~\ref{tab_ab_pathway}, both the two modules demonstrate strong video modeling capability.
When the ResNet baseline is enhanced with the EABs, the Top1 accuracy is significantly improved by 2.2\%. The reason is that the features extracted by the ResNet baseline are not accurate enough, and the proposed EABs can refine the features with the dynamic spatial-temporal kernel.
For a more intuitive understanding, we will visualize
the refined feature maps by our method in section~\ref{sec_eab}.
Then, we observe that the ResNet + SOI-Tr baseline also outperforms the original ResNet baseline by 0.7\% in terms of Top1 accuracy, while only introducing an extra 0.7 GFLOPs computation cost.
Finally, simultaneously using EAB and SOI-Tr boosts the performance to 51.9\%, which proves the complementarity of the two proposed modules.

\begin{figure}[!htb]
	\centering
	\begin{minipage}[t]{0.76\linewidth}
		\centering
		\centerline{\includegraphics[width=8.0cm]{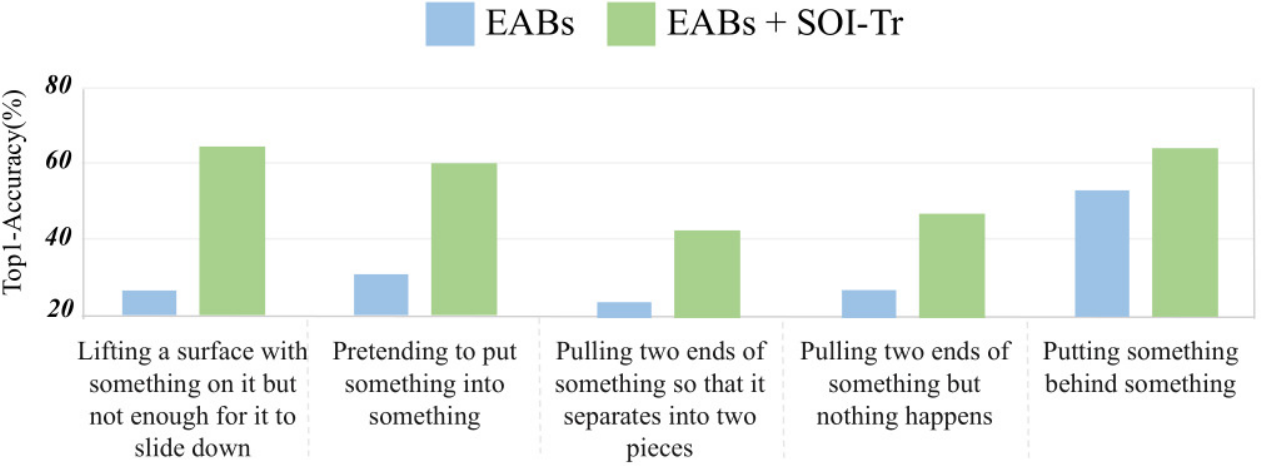}}
	\end{minipage}
	\caption {
		The top 5 action classes that are significantly improved after introducing the SOI-Tr module.
	}
	\label{fig_ab_pathway_comp}
\end{figure}

\begin{figure}[!thp]
	\centering
	\begin{minipage}[t]{0.74\linewidth}
		\centering
		\centerline{\includegraphics[width=8.2cm]{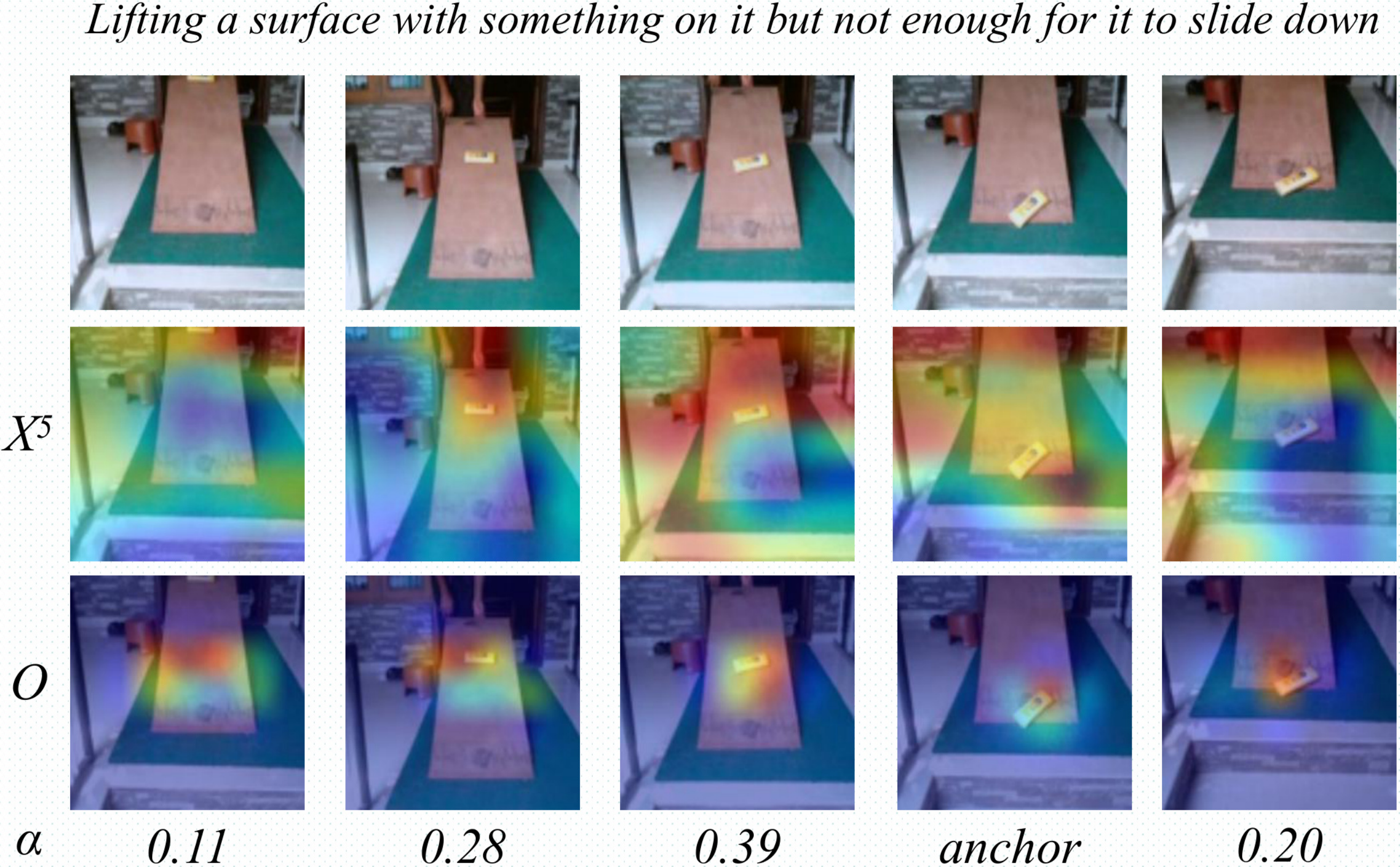}}
	\end{minipage}
	\caption {
		Visualization of one video clip from the most improved category by the SOI-Tr.
		The input clip is first processed by EABs to obtain the spatial-temporal feature map {$X^5$}.
		Then, SOI-Tr calculates the saliency map {$O$} of the most concentrated object and the interactions of this object across the temporal axis. We take the 4th frame as the anchor and show the attention vector {$\alpha$}.
	}
	\spaceabovetab
	\label{fig_vis_ean_soitr}
\end{figure}
We also plot the top 5 classes that are significantly improved after introducing the SOI-Tr. As shown in Fig.~\ref{fig_ab_pathway_comp},
we find that the most improved instances can be roughly divided into two groups:
{(a)} The instances that require tracking the state of a certain object over the whole clip, such as the videos of \textit{``Lifting a surface ... ''} and \textit{``Pulling two ends ... ''}.
{(b)} The instances that contain multiple objects and the interactions between them, such as the videos of \textit{``Pretending to put ... ''}.
This is aligned with the motivation of introducing SOI-Tr, \textit{i.e.}, accurately modeling the long-range object interactions benefits the recognition of some complex actions.

To systematically understand how the EABs and SOI-Tr improve the recognition performance, we randomly select one video from the category \textit{``Lifting a surface with something on it but not enough for it to slide down''} and visualize it.
In Fig.~\ref{fig_vis_ean_soitr}, we can clearly see that the original feature $X^5$ before global modeling concentrates on the background or the board, omitting the main object, \textit{i.e.}, the small sliding box. This makes sense because both the spatial area and the motion magnitude of the board are more obvious than the small box. After introducing the SOI-Tr, the object detector first finds the main object. Then, the Transformer model builds the long-range dependencies across the whole clip. We also notice that the board in the first frame is also detected. But, this background object will be neglected in the self-attention model because its weight is only 0.11.

\begin{table}[!thbp]
		\caption{Study on the location of EAB and SOI-Tr.	}
	\centering
	\setlength{\tabcolsep}{1.5mm}
	\renewcommand{\arraystretch}{1.0}
	\begin{tabular}{lccccc}
		\Xhline{2\arrayrulewidth}
		\multirow{2}{*}{\tabincell{c}{\textbf{EAB} } } &\multirow{2}{*}{\tabincell{c}{\textbf{SOI-Tr}} } 
		& \multirow{2}{*}{\tabincell{c}{\textbf{Param}} }
		&\multirow{2}{*}{\tabincell{c}{\textbf{FLOPs}} } & \multicolumn{2}{c}{\textbf{Something V1}}  \\
		&&&& \scriptsize{\textbf{Top1 (\%)}} & \scriptsize{\textbf{Top5 (\%)}} \\ 
		\hline
		Stage 1$\sim$2 & Stage 3$\sim$5 &35.6M & 35.8G & 49.4 & 78.2\\
		Stage 1$\sim$3 & Stage 4$\sim$5 &34.8M & 35.9G & 50.4 & 79.2\\
		Stage 1$\sim$4 & Stage 5 		&36.0M & 36.1G & \textbf{51.9} & \textbf{79.5}\\
		Stage 1$\sim$5 & - 		 		&65.8M & 36.2G & 50.8 & 79.1\\
		\Xhline{2\arrayrulewidth}
		
	\end{tabular}

	\label{tab_insert_blk}
\end{table}

\textit{\textbf{Q2: Where to insert the proposed modules?}}
We perform an ablation study on which stage to use local operator (EAB) and global operator (SOI-Tr). The results are shown in Tab.~\ref{tab_insert_blk}.
From these results, we see that adding more EABs into the main network only slightly increases the computational cost due to the high efficiency of the bottleneck designing and group convolution. When some EABs are replaced with the SOI-Tr, the performance decreases consistently. This implies that the local spatial-temporal information is crucial for action recognition, which cannot be substituted by the high-level object interaction information. We also try to build the network only with EABs, the result is also inferior to the original hybrid model (convolution+self-attention). The setting of using EAB after stage 1$\sim$4 and SOI-Tr after stage 5 obtains the best recognition accuracy and is with reasonable complexity.

\textit{\textbf{Q3: Is the prior assumption of SOI-Tr reasonable?}}
To prove the end-to-end foreground object detector and the sparsity assumption for object interactions are both important for SOI-Tr, we train other variant models where we replace our detected object regions with the same number of the fixed regions or the regions detected by a pre-trained Faster RCNN~\cite{ren2016faster} model.
When the number of the boxes output from Faster RCNN is too small, we pad it with the central region of the frames.
The performances of the models are compared in Tab.~\ref{tab_interact_models}.

\begin{table}[!htbp]
	\vspace{-3mm}
	\caption{
		Study on various object region sampling strategies.
	}
	\centering
	\setlength{\tabcolsep}{4.5mm}
	\renewcommand{\arraystretch}{1.0}
	\begin{tabular}{lccc}
		\Xhline{2\arrayrulewidth}
		\multirow{2}{*}{\tabincell{c}{\textbf{Regions} } }
		&\multirow{2}{*}{\tabincell{c}{\textbf{FLOPs}} } & \multicolumn{2}{c}{\textbf{Something V1}}  \\
		&& \scriptsize{\textbf{Top1 (\%)}} & \scriptsize{\textbf{Top5 (\%)}} \\ 
		\hline
		None & 35.3G & 50.8 & 78.4 \\
		\hline
		Fixed & 35.8G & 50.9 & 78.4 \\
		Faster RCNN & 71.3G & 51.1 & 78.7 \\
		All & 36.4G & 51.3 & 79.1 \\
		Our Det-Net & 36.1G & \textbf{51.9} & \textbf{79.5} \\
		\Xhline{2\arrayrulewidth}
	\end{tabular}

	\label{tab_interact_models}
	\spacebelowtab
\end{table}

 First, we notice that building the interaction model upon the fixed regions already slightly improves the performance, proving that the interaction modeling is beneficial to the action recognition. Then we use the Faster RCNN detector to predict more accurate foreground regions. Surprisingly, the performance improvement is negligible. This may be ascribed to the fact that most frames only contain one or two objects, which cover fewer regions compared with the ``fixed region'' scheme. In contrast, the Saliency-Net embedded in our method always detects enough salient regions in an end-to-end manner and obtains the best performance, \textit{i.e.}, 51.9\%. Also, it is computationally efficient due to the shared feature extractor with the other parts of the framework. We emphasize that our embedded Saliency-Net outperforms Faster RCNN by 0.8\% while running 118$\times$ faster.
 
 We further try to leverage all positions to build a dense interaction model, as shown in the penultimate row of Tab.~\ref{tab_interact_models}. However, the performance is obviously inferior to our method. This strongly supports our assumption that most regions are only background noises for the final prediction and leveraging all of them will deteriorate the final performance.

\subsection{Further Studies for EAB}\label{sec_eab}
In this section, we make further studies on the aspects that impact the effectiveness of EAB.

\textbf{Large receptive field and multi-scale modeling are important.}
To verify this, we introduce the following baselines: 

\begin{figure}[!htb]
	\centering
	\begin{minipage}[b]{0.9\linewidth}
		\centering
		\centerline{\includegraphics[width=8cm]{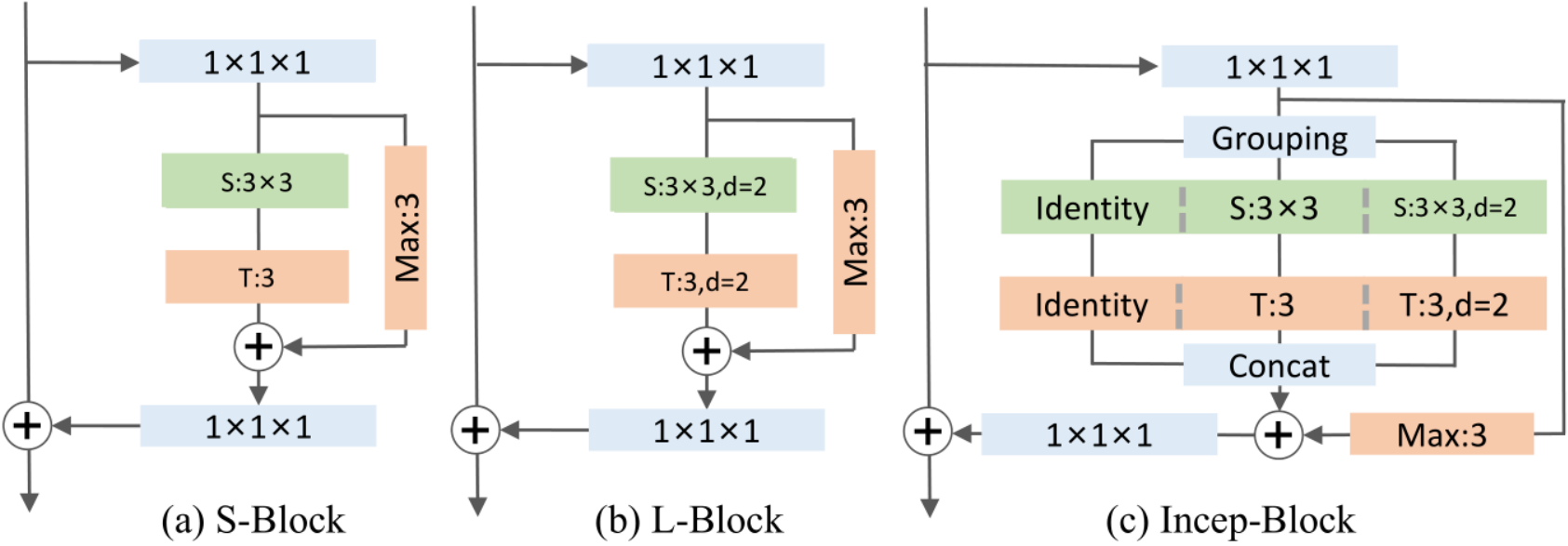}}
	\end{minipage}
	\caption {Illustrations of the baseline spatial-temporal blocks.
		The representation signs are the same meaning as Fig.~\ref{fig_ean}.
	}
\spacebelowtab
	\label{fig_base_arch}
\end{figure}

	\noindent {\textsf{\small (1) S-Block}}.
	It is implemented with a (2+1)D convolution with kernel size $3\times3\times3$ and group size 3, as shown in Fig.~\ref{fig_base_arch} (a), which only captures single-scale features with a small receptive field.
	
	\noindent \textsf{\small (2) L-Block}.
	It is implemented with a (2+1)D convolution with kernel size $3\times3\times3$, group size 3, and dilation size $2\times2\times2$, as shown in Fig.~\ref{fig_base_arch} (b), which captures single-scale features with a larger receptive field.
	
	\noindent \textsf{\small (3) Incep-Block}.
	It is implemented with a group of (2+1)D convolutions in an Inception-style, as shown in Fig.~\ref{fig_base_arch} (c), which captures multi-scale features with a larger receptive field.
	 The only difference between this baseline and EAB is that the $M$ is replaced with an identity mapping operation.

\begin{table}[!htbp]
		\caption{
		Comparison of different spatial-temporal blocks. RFS denotes the receptive field size.
	}
	\centering
	\setlength{\tabcolsep}{0.64mm}
	\renewcommand{\arraystretch}{1.0}
	\begin{tabular}{lccccccc}
		\Xhline{2\arrayrulewidth}
		\multirow{2}{*}{\tabincell{c}{\textbf{Models}} } & 
		\multirow{2}{*}{\textbf{\tabincell{c}{RFS}}}&
		
		\multirow{2}{*}{\textbf{\tabincell{c}{Multi\\scale?}}} & 
		\multirow{2}{*}{\tabincell{c}{\textbf{Param}}} & 
		\multirow{2}{*}{\tabincell{c}{\textbf{FLOPs}}} & 
		\multicolumn{2}{c}{\textbf{Something V1}}  \\
		&&&&& \scriptsize{\textbf{Top1 (\%)}} & \scriptsize{\textbf{Top5 (\%)}} \\ 
		\hline
		Only SOI-Tr &-&-&30.3M& 33.8G &49.3 & 77.9 \\
		\hline
		+S-Block &$3\times3\times3$&-&30.9M &36.1G   & 49.6 & 78.1 \\
		+L-Block &$5\times5\times5$&-&30.9M &36.1G   & 50.3 & 78.4 \\
		+Incep-Block &$5\times5\times5$&\checkmark &30.9M & 36.0G & 50.8 & 78.8 \\
		\hline
		\textbf{+EAB} &\textbf{$5\times5\times5$} &\checkmark &36.0M &36.1G  &\textbf{51.9} & \textbf{79.5}   \\
		\Xhline{2\arrayrulewidth}
	\end{tabular}
	\label{tab_st_model}
\end{table}

We compare our method with the proposed baseline methods in Tab.~\ref{tab_st_model}. First, it can be seen that EAB outperforms the S-Block baseline by a large margin (51.9\% \textit{vs.} 49.6\%). The improvement is originated from two aspects: (1) The large spatial-temporal kernel within EAB enables the larger receptive field and aggregates more local information. (2) The explicit multi-scale modeling introduces richer feature representation.
It is necessary to validate the independent contribution from the two aspects. We first compare the S-Block baseline with the L-Block baseline. L-Block has a larger spatial-temporal receptive field size but the same number of parameters. We can see that the recognition accuracy is improved by 0.7\%.
Then, we build the Incep-Block baseline by enhancing the L-Block baseline with multi-scale modeling capability. This improvement further improves the recognition accuracy. From the comparisons above, we verify that both the two aspects facilitate the action task, and multi-scale architecture fully exploits the large receptive field.

\begin{figure}[!thbp]
	\centering
	\centerline{\includegraphics[width=8.2cm]{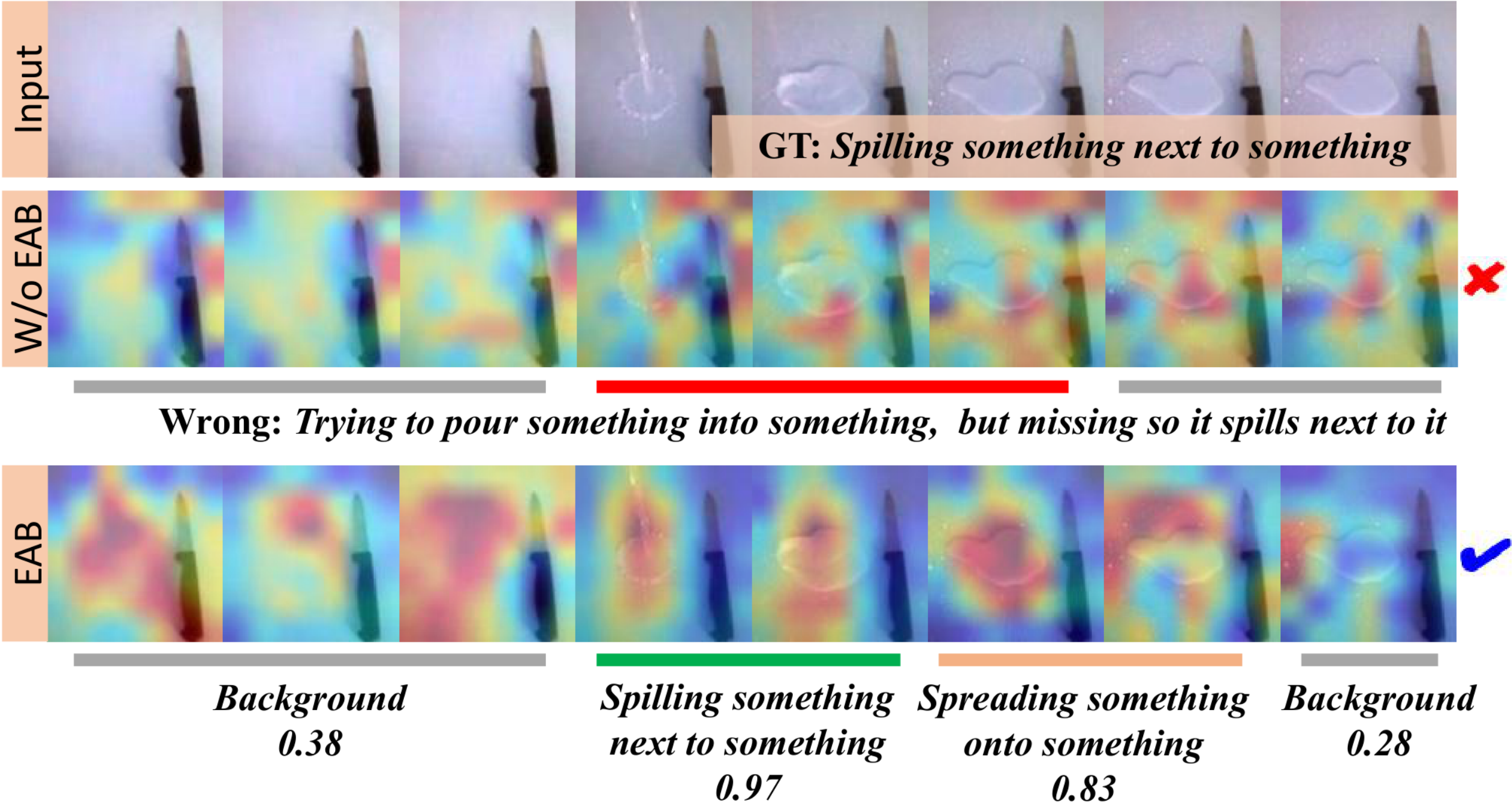}}
	\caption {Comparison of the features from Stage4 of EAN and the evolution of the prediction scores. The proposed EAB can discover more semantically-aligned regions for actions, and also suppress the noisy information from backgrounds to yield correct prediction. \textit{Zoom in for better visualization.}
	}
	\label{fig_vis_feature}
\end{figure}

We randomly select one video from Something V1 dataset and visualize the $8\times14\times14$ feature map output from Stage4 (this is before the inserting of the SOI-Tr), as shown in Fig.~\ref{fig_vis_feature}. It clearly demonstrates that our method can discover more semantically consistent regions for actions, and in the meantime reduce noisy backgrounds for correct prediction. Moreover, the feature activation heatmaps of our method are better spatially aligned with the target object (see the water).
We also show the state evolution process in Fig.~\ref{fig_vis_feature}. Interestingly, our method detects the start and end points of actions although only trained with classification labels.

\textbf{Dynamic architecture matters.}
From Tab.~\ref{tab_st_model}, we notice that the performance gap between the Incep-Block baseline and EAB is still rather large, \textit{i.e.}, 1.1\% Top1 accuracy. We conjecture this is due to the dynamic architecture of EAB. As mentioned in section~\ref{EAB}, the inference pathway for EAB is determined by the kernel fusion matrix $M$. For more detailed analysis, we propose two kernel fusion strategies:

\noindent {\textsf{\small (1) Channel Shuffle}}. We replace $M$ with a conventional fusion method, \textit{i.e.}, Channel Shuffle operation~\cite{zhang2018shufflenet}, which enables the communication of the features of different groups.

\noindent {\textsf{\small (2) Static Matrix.}} The $M$ is a learnable matrix during training. But, it's a fixed matrix during inference.

\begin{table}[!h]
	\caption{
		Comparison of different feature fusion methods.
	}
	\centering
	\small
	\setlength{\tabcolsep}{1.8mm}
	\renewcommand{\arraystretch}{1.0}
	\begin{tabular}{lccccc}
		\Xhline{2\arrayrulewidth}
		\multirow{2}{*}{\tabincell{c}{ \textbf{Methods} } } &
		\multirow{2}{*}{\tabincell{c}{\textbf{Param}  } } & 
		\multirow{2}{*}{\tabincell{c}{\textbf{FLOPs}  } } & 
		\multicolumn{2}{c}{\textbf{Something V1}}  \\
		
		&&& \scriptsize{\textbf{Top1 (\%)}} & \scriptsize{\textbf{Top5 (\%)}} \\ 
		\hline
		Identity (Incep-Block) & 30.9M &36.0G & 50.8 & 78.8 \\
		\hline
		Channel Shuffle & 30.9M &35.9G & 51.1 & 78.9 \\
		Static Matrix & 30.9M &36.0G & 51.3 & 79.2 \\
		Dynamic Matrix & 36.0M & 36.1G & \textbf{51.9} & \textbf{79.5} \\
		\Xhline{2\arrayrulewidth}
	\end{tabular}
	
	\label{tab_ft_inter}
\end{table}

Both the above two baselines belong to the static architecture but they are similar to the EAB in terms of the network details, which are perfect for studying the impact of dynamic modeling. We compare their performances in Tab.~\ref{tab_ft_inter}. We observe that the performance improves consistently with a more complex kernel fusion strategy, \textit{i.e.}, Identity $\rightarrow$Channel Shuffle $\rightarrow$Static Matrix $\rightarrow$Dynamic Matrix.
The Dynamic fusion matrix adopted by EAB shows the best performance (51.9\%) with negligible extra cost.

\begin{figure}[!t]
	\centering
	\centerline{\includegraphics[width=8cm]{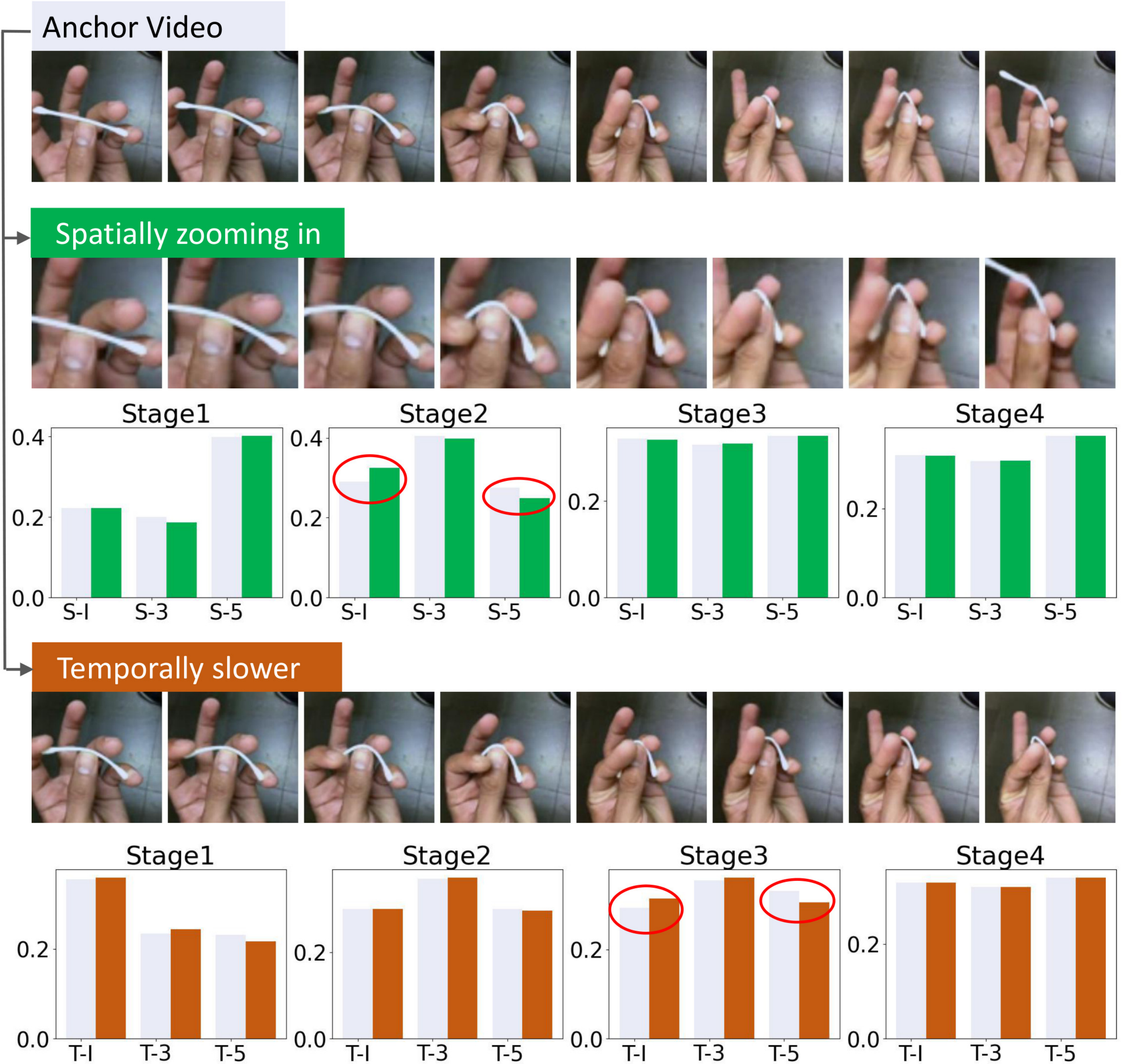}}
	\caption {Visualizing the dynamic kernel fusion matrix $M$ of the proposed EAB via the kernel weights. For each spatial or temporal kernel, its weight is computed by summing the matrix values connected to it.
		In the first row, we give an anchor video.
		Below it, we show the impact to the kernel weights by changing the spatial or temporal scales of the anchor video. The kernel weights of anchor, spatially-, and temporally-augmented videos are indicated by \textcolor{gray}{gray}, \textcolor{green}{green}, and \textcolor{brown}{brown} bars, respectively.
		``S-3''\&``T-3'' denotes the fixed-size spatial\&temporal kernel of size 3$\times$3\&3.
	}
	\label{fig_m}
\end{figure}

\textbf{Kernel visualization.}
To verify that the dynamic kernel fusion matrix $M$ of EAB is indeed adaptive to the scales of the main objects and the key events within different videos, we conduct a group of experiments by augmenting one anchor video and observing the change of the weights of the fixed-scale kernels. The augmented videos and the kernel weight changing procedure are illustrated in Fig.~\ref{fig_m}. We first see, both the weight distributions along the temporal axis or the spatial axis are not sparse, \textit{i.e.}, all kernels are activated. This supports our assumption that the optimal spatial-temporal kernel for the video is with an unknown complex shape and cannot be accurately replaced by one kernel of fixed-scale. Also, the distributions do not follow some simple distributions such as Uniform or Gaussian, indicating that the kernel weights cannot be trivially hand-crafted and are required to be learned from data.
When we spatially zoom in the anchor video by 1.6$\times$, the main objects in the video, \textit{i.e.}, the hand and the stick, are easier to be discovered, we see that EAB is more inclined to exploit the spatial convolutions of small kernels such as that of size 1$\times$1 instead of that of size 5$\times$5
. Similarly, when we sample the frames with 2$\times$ higher frame-rate, the object motions become slower and the small temporal convolutions such as that with kernel size 1 are fully used.

\begin{table}[!thbp]
	\caption{
		Ablation on detailed designs of EAB.
	}
	\centering
	\setlength{\tabcolsep}{2.0mm}
	\renewcommand{\arraystretch}{1.0}
	\begin{tabular}{lcccc}
		\Xhline{2\arrayrulewidth}
		\multirow{2}{*}{\tabincell{c}{\textbf{Design} } } &
		\multirow{2}{*}{\tabincell{c}{\textbf{Param} } } &
		\multirow{2}{*}{\tabincell{c}{\textbf{FLOPs} } } &
		\multicolumn{2}{c}{\textbf{Something V1}}  \\
		&&& \scriptsize{\textbf{Top1 (\%)}} & \scriptsize{\textbf{Top5 (\%)}} \\ 
		\hline
		Without Max Pool &{36.0M}& {36.1G}& {50.6} & {78.4}   \\
		Avg Pool &{36.0M}& {36.1G}& {51.4} & {79.8}   \\
		Without inter ReLU &{36.0M}& {36.1G}& {50.9} & {79.0}   \\
		Without dilation &{37.2M}& {37.5G} & \textbf{51.9} & \textbf{79.8}   \\
		(1+1+1)D  &{35.7M}& {35.7G}& {51.2} & {79.2}   \\
		\hline
		Ours &{36.0M}& {36.1G}& \textbf{51.9} & {79.5}   \\
		\Xhline{2\arrayrulewidth}
	\end{tabular}
	
	\label{tab_arch_details}
\end{table}

\textbf{\textbf{Studies on EAB details.}}
In this part, we conduct experiments to verify whether all the designs of EAB contribute to the final performance.
As shown in Tab.~\ref{tab_arch_details},
the max pooling operation significantly improves the performance ({1.3\%} \textit{w.r.t} Top1 accuracy), and meanwhile our method is not sensitive to specific implementation of this operation. 
Both average pooling and max pooling operators achieve excellent performance.
Max pooling demonstrates a slight advantage over average pooling because the regions of the key objects and frames related to action only cover a small proportion of the input video data.
Also, we find that the extra non-linearity introduced by the intermediate ReLU operations between spatial- and temporal-filters also benefit the performance, which is consistent with the conclusion from previous work~\cite{tran2018closer}.
Besides, we demonstrate that the dilated convolution achieves comparable performance with the ordinary convolution while it is much more efficient.
Finally, we also try to decompose the 2D spatial convolution into two stacking 1D convolutions. But this brings a slight performance drop. To summarize, the extensive experiments in this section prove the necessity of detailed designs in EAB.

\section{Erroneous Cases and Limitations}

Although the quantitative results on standard benchmarks and the extensive analysis above have verified the effectiveness of the proposed framework, it inevitably has some limitations, which lead to erroneous recognition results.

\begin{figure}[!thp]
	\centering
	\centerline{\includegraphics[width=8.2cm]{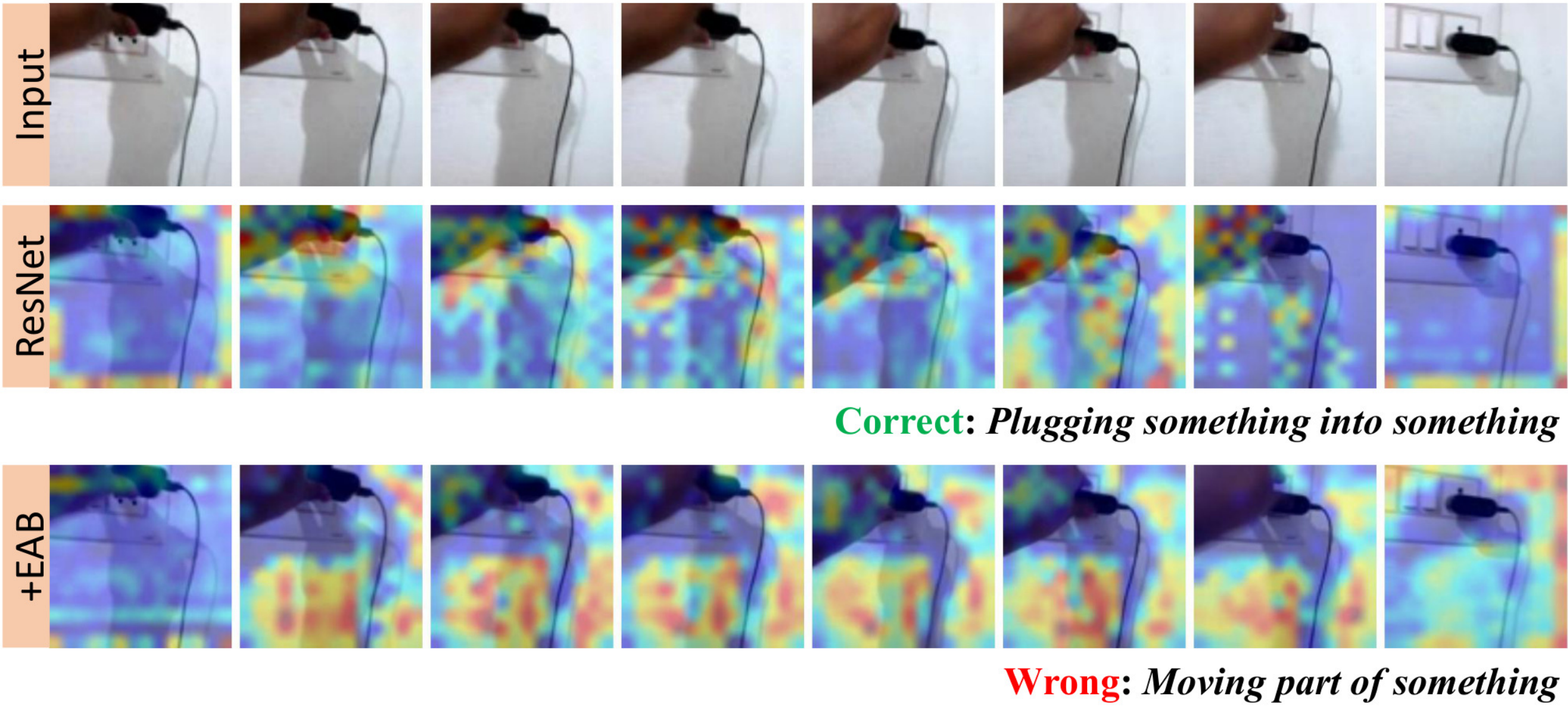}}
	\caption {
		Recognition error caused by EAB.
	The shadow of the human hand instead of the real hand and the charger is attended, causing the recognition result changing from \textit{plugging something into something} to \textit{moving part of something}.
	}
	\label{fig_error_eab}
\end{figure}

One limitation is caused by the simple architecture of ESP-Net within EAB. ESP-Net is responsible for perceiving the event scales within the input video, composed of two convolution layers followed by a global average pooling operation. Although this simple ``average'' operation is lightweight in terms of the computational cost, it also makes the statistical results of the video biased to the large objects.
As shown in Fig.~\ref{fig_error_eab}, the feature activations are dominated by the large-area human hand shadow, neglecting the real objects (the human hand and the charger) involved in the action \textit{plugging something into something}.

\begin{figure}[!thp]
	\centering
	\centerline{\includegraphics[width=8.2cm]{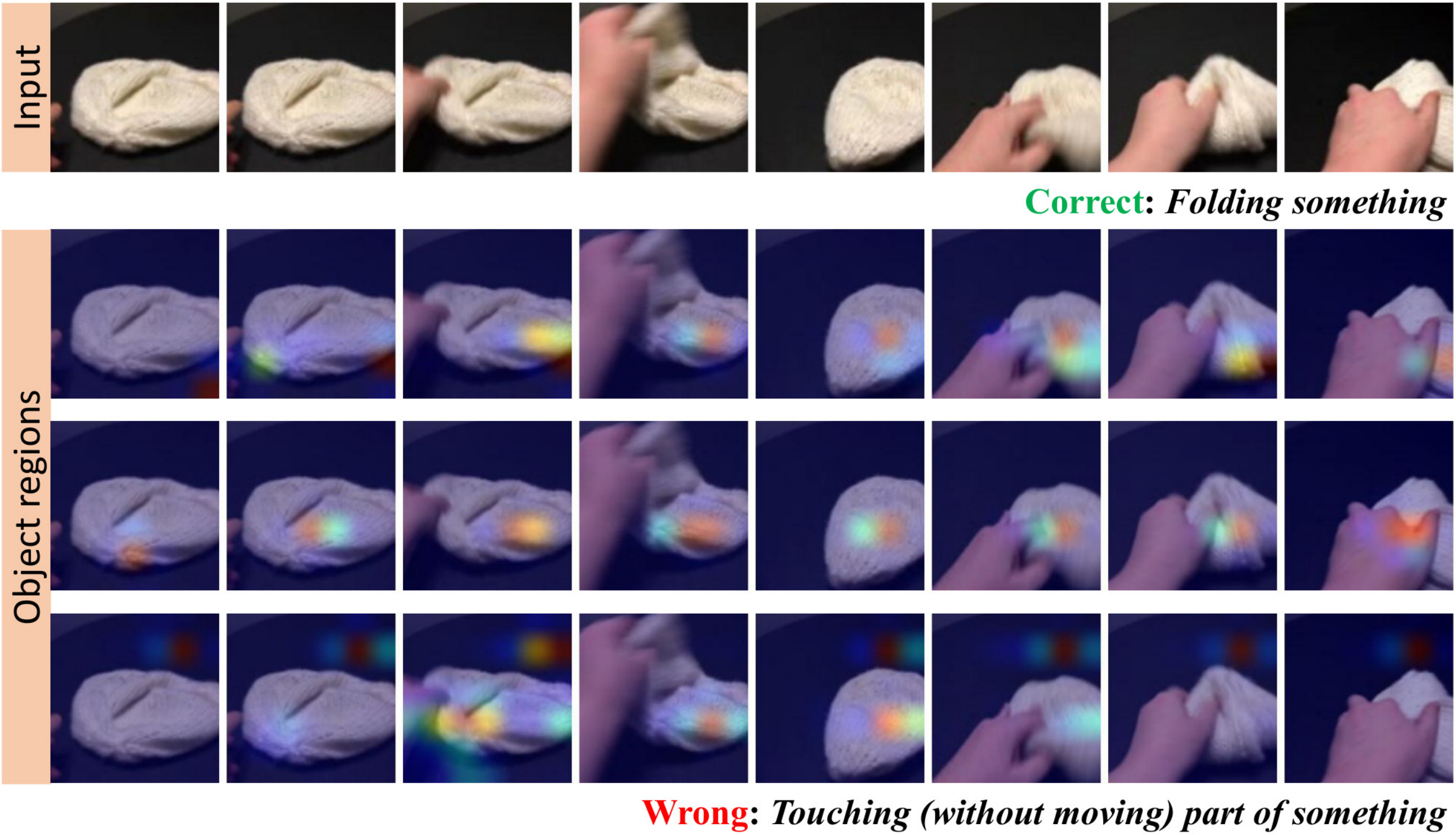}}
	\caption{
		Recognition error caused by SOI-Tr.
		We visualize the foreground object distribution maps, where only parts of the {towel} are detected in this case.
		Therefore, the global state \textit{folding} of the {towel} can not be perceived, resulting the biased action recognition result \textit{touching}.
	}
	\label{fig_error_soitr}
\end{figure}

Another limitation is originated from the proposed SOI-Tr.
The adaptiveness of SOI-Tr lies in detecting different foreground objects for different input videos.
Nevertheless, the adaptiveness may be limited by the representation of the objects, \textit{i.e.}, points in the feature map, which correspond to fixed-size regions within the input video. Therefore, the granularity and the scale of the detected foreground objects are not flexible enough.
As shown in Fig.~\ref{fig_error_soitr}, only small parts of the towel can be detected.
Therefore, the global state \textit{folding} of the towel can not be perceived. Instead, the local states of the wrongly attended objects, \textit{i.e.}, the human hand and the partial towel, contribute to the wrong prediction \textit{touching part of something}.

\section{Conclusion and Future Works}
To model the spatial-temporal scale variances and the long-range object interactions in videos, we propose to dynamically generate the video-adaptive kernels from the input video and model the interactions among the objects with a Transformer. Moreover, we design a novel short-term motion representation to further enhance the performance of our method. We perform extensive evaluations to study the effectiveness of the proposed approach on video action recognition task, and the results demonstrate that our models achieve impressive performances on Something-Something V1/V2, Kinetics-400, and Diving48 datasets.
In the future, we will explore how to better approximate the video-adaptive kernel. As for the network architecture, we will investigate more powerful backbone networks. We also plan to extend the proposed framework to more downstream video tasks such as the spatial-temporal action localization task. 

\begin{acknowledgements}
	This work was supported by NSFC (61831015), National Key R\&D Program of China (2021YFE0206700), ~~NSFC (U19B2035), Shanghai Municipal Science and Technology Major Project (2021SHZDZX0102), and CAAI-Huawei MindSpore Open Fund.
\end{acknowledgements}

\bibliographystyle{spmpsci}      

{\footnotesize
\bibliography{mybib}

\begin{thebibliography}{10}
\providecommand{\url}[1]{{#1}}
\providecommand{\urlprefix}{URL }
\expandafter\ifx\csname urlstyle\endcsname\relax
  \providecommand{\doi}[1]{DOI~\discretionary{}{}{}#1}\else
  \providecommand{\doi}{DOI~\discretionary{}{}{}\begingroup
  \urlstyle{rm}\Url}\fi

\bibitem{arnab2021vivit}
Arnab, A., Dehghani, M., Heigold, G., Sun, C., Lu{\v{c}}i{\'c}, M., Schmid, C.:
  Vivit: A video vision transformer.
\newblock In: Proceedings of the IEEE/CVF International Conference on Computer
  Vision, pp. 6836--6846 (2021)

\bibitem{bensch2017spatiotemporal}
Bensch, R., Scherf, N., Huisken, J., Brox, T., Ronneberger, O.: Spatiotemporal
  deformable prototypes for motion anomaly detection.
\newblock International Journal of Computer Vision \textbf{122}(3), 502--523
  (2017)

\bibitem{bertasius2018learning}
Bertasius, G., Feichtenhofer, C., Tran, D., Shi, J., Torresani, L.: Learning
  discriminative motion features through detection.
\newblock arXiv preprint arXiv:1812.04172  (2018)

\bibitem{bertasius2021space}
Bertasius, G., Wang, H., Torresani, L.: Is space-time attention all you need
  for video understanding?
\newblock arXiv preprint arXiv:2102.05095  (2021)

\bibitem{bulat2021space}
Bulat, A., Perez~Rua, J.M., Sudhakaran, S., Martinez, B., Tzimiropoulos, G.:
  Space-time mixing attention for video transformer.
\newblock Advances in Neural Information Processing Systems \textbf{34} (2021)

\bibitem{carreira2017quo}
Carreira, J., Zisserman, A.: Quo vadis, action recognition? a new model and the
  kinetics dataset.
\newblock In: proceedings of the IEEE Conference on Computer Vision and Pattern
  Recognition, pp. 6299--6308 (2017)

\bibitem{chen2021sportscap}
Chen, X., Pang, A., Yang, W., Ma, Y., Xu, L., Yu, J.: Sportscap: Monocular 3d
  human motion capture and fine-grained understanding in challenging sports
  videos.
\newblock International Journal of Computer Vision \textbf{129}(10), 2846--2864
  (2021)

\bibitem{chen2020dynamic}
Chen, Y., Dai, X., Liu, M., Chen, D., Yuan, L., Liu, Z.: Dynamic convolution:
  Attention over convolution kernels.
\newblock In: Proceedings of the IEEE/CVF Conference on Computer Vision and
  Pattern Recognition, pp. 11,030--11,039 (2020)

\bibitem{cherian2019second}
Cherian, A., Gould, S.: Second-order temporal pooling for action recognition.
\newblock International Journal of Computer Vision \textbf{127}(4), 340--362
  (2019)

\bibitem{cong2021spatial}
Cong, Y., Liao, W., Ackermann, H., Rosenhahn, B., Yang, M.Y.: Spatial-temporal
  transformer for dynamic scene graph generation.
\newblock In: Proceedings of the IEEE/CVF International Conference on Computer
  Vision, pp. 16,372--16,382 (2021)

\bibitem{deng2009imagenet}
Deng, J., Dong, W., Socher, R., Li, L.J., Li, K., Fei-Fei, L.: Imagenet: A
  large-scale hierarchical image database.
\newblock In: 2009 IEEE conference on computer vision and pattern recognition,
  pp. 248--255. Ieee (2009)

\bibitem{dosovitskiy2020image}
Dosovitskiy, A., Beyer, L., Kolesnikov, A., Weissenborn, D., Zhai, X.,
  Unterthiner, T., Dehghani, M., Minderer, M., Heigold, G., Gelly, S., et~al.:
  An image is worth 16x16 words: Transformers for image recognition at scale.
\newblock arXiv preprint arXiv:2010.11929  (2020)

\bibitem{fan2021multiscale}
Fan, H., Xiong, B., Mangalam, K., Li, Y., Yan, Z., Malik, J., Feichtenhofer,
  C.: Multiscale vision transformers.
\newblock arXiv preprint arXiv:2104.11227  (2021)

\bibitem{feichtenhofer2019slowfast}
Feichtenhofer, C., Fan, H., Malik, J., He, K.: Slowfast networks for video
  recognition.
\newblock In: Proceedings of the IEEE international conference on computer
  vision, pp. 6202--6211 (2019)

\bibitem{feichtenhofer2020deep}
Feichtenhofer, C., Pinz, A., Wildes, R.P., Zisserman, A.: Deep insights into
  convolutional networks for video recognition.
\newblock International Journal of Computer Vision \textbf{128}(2), 420--437
  (2020)

\bibitem{feichtenhofer2016convolutional}
Feichtenhofer, C., Pinz, A., Zisserman, A.: Convolutional two-stream network
  fusion for video action recognition.
\newblock In: Proceedings of the IEEE conference on computer vision and pattern
  recognition, pp. 1933--1941 (2016)

\bibitem{ferryman2000visual}
Ferryman, J.M., Maybank, S.J., Worrall, A.D.: Visual surveillance for moving
  vehicles.
\newblock International Journal of Computer Vision \textbf{37}(2), 187--197
  (2000)

\bibitem{gao2019res2net}
Gao, S.H., Cheng, M.M., Zhao, K., Zhang, X.Y., Yang, M.H., Torr, P.: Res2net: A
  new multi-scale backbone architecture.
\newblock IEEE transactions on pattern analysis and machine intelligence
  \textbf{43}(2), 652--662 (2019)

\bibitem{girdhar2019video}
Girdhar, R., Carreira, J., Doersch, C., Zisserman, A.: Video action transformer
  network.
\newblock In: Proceedings of the IEEE Conference on Computer Vision and Pattern
  Recognition, pp. 244--253 (2019)

\bibitem{girdhar2021anticipative}
Girdhar, R., Grauman, K.: Anticipative video transformer.
\newblock In: Proceedings of the IEEE/CVF International Conference on Computer
  Vision, pp. 13,505--13,515 (2021)

\bibitem{goyal2017something}
Goyal, R., Kahou, S.E., Michalski, V., Materzynska, J., Westphal, S., Kim, H.,
  Haenel, V., Fruend, I., Yianilos, P., Mueller-Freitag, M., et~al.: The"
  something something" video database for learning and evaluating visual common
  sense.
\newblock In: Proceedings of the IEEE international conference on computer
  vision, vol.~1, p.~5 (2017)

\bibitem{hara2017learning}
Hara, K., Kataoka, H., Satoh, Y.: Learning spatio-temporal features with 3d
  residual networks for action recognition.
\newblock In: Proceedings of the IEEE International Conference on Computer
  Vision Workshops, pp. 3154--3160 (2017)

\bibitem{he2016deep}
He, K., Zhang, X., Ren, S., Sun, J.: Deep residual learning for image
  recognition.
\newblock In: Proceedings of the IEEE conference on computer vision and pattern
  recognition, pp. 770--778 (2016)

\bibitem{ilg2017flownet}
Ilg, E., Mayer, N., Saikia, T., Keuper, M., Dosovitskiy, A., Brox, T.: Flownet
  2.0: Evolution of optical flow estimation with deep networks.
\newblock In: Proceedings of the IEEE conference on computer vision and pattern
  recognition, pp. 2462--2470 (2017)

\bibitem{jia2016dynamic}
Jia, X., De~Brabandere, B., Tuytelaars, T., Gool, L.V.: Dynamic filter
  networks.
\newblock Advances in neural information processing systems \textbf{29},
  667--675 (2016)

\bibitem{jiang2019stm}
Jiang, B., Wang, M., Gan, W., Wu, W., Yan, J.: Stm: Spatiotemporal and motion
  encoding for action recognition.
\newblock In: Proceedings of the IEEE International Conference on Computer
  Vision, pp. 2000--2009 (2019)

\bibitem{kanojia2019attentive}
Kanojia, G., Kumawat, S., Raman, S.: Attentive spatio-temporal representation
  learning for diving classification.
\newblock In: Proceedings of the IEEE Conference on Computer Vision and Pattern
  Recognition Workshops (2019)

\bibitem{khowaja2020semantic}
Khowaja, S.A., Lee, S.L.: Semantic image networks for human action recognition.
\newblock International Journal of Computer Vision  (2020)

\bibitem{kwon2020motionsqueeze}
Kwon, H., Kim, M., Kwak, S., Cho, M.: Motionsqueeze: Neural motion feature
  learning for video understanding.
\newblock In: European Conference on Computer Vision, pp. 345--362. Springer
  (2020)

\bibitem{li2020tea}
Li, Y., Ji, B., Shi, X., Zhang, J., Kang, B., Wang, L.: Tea: Temporal
  excitation and aggregation for action recognition.
\newblock In: Proceedings of the IEEE/CVF Conference on Computer Vision and
  Pattern Recognition, pp. 909--918 (2020)

\bibitem{li2018resound}
Li, Y., Li, Y., Vasconcelos, N.: Resound: Towards action recognition without
  representation bias.
\newblock In: Proceedings of the European Conference on Computer Vision, pp.
  513--528 (2018)

\bibitem{lin2019tsm}
Lin, J., Gan, C., Han, S.: Tsm: Temporal shift module for efficient video
  understanding.
\newblock In: Proceedings of the IEEE International Conference on Computer
  Vision, pp. 7083--7093 (2019)

\bibitem{liu2020teinet}
Liu, Z., Luo, D., Wang, Y., Wang, L., Tai, Y., Wang, C., Li, J., Huang, F., Lu,
  T.: Teinet: Towards an efficient architecture for video recognition.
\newblock In: Proceedings of the AAAI Conference on Artificial Intelligence,
  vol.~34, pp. 11,669--11,676 (2020)

\bibitem{liu2020tam}
Liu, Z., Wang, L., Wu, W., Qian, C., Lu, T.: Tam: Temporal adaptive module for
  video recognition.
\newblock arXiv preprint arXiv:2005.06803  (2020)

\bibitem{lu2019fast}
Lu, C., Shi, J., Wang, W., Jia, J.: Fast abnormal event detection.
\newblock International Journal of Computer Vision \textbf{127}(8), 993--1011
  (2019)

\bibitem{luo2019grouped}
Luo, C., Yuille, A.L.: Grouped spatial-temporal aggregation for efficient
  action recognition.
\newblock In: Proceedings of the IEEE International Conference on Computer
  Vision, pp. 5512--5521 (2019)

\bibitem{ma2018attend}
Ma, C.Y., Kadav, A., Melvin, I., Kira, Z., AlRegib, G., Peter~Graf, H.: Attend
  and interact: Higher-order object interactions for video understanding.
\newblock In: Proceedings of the IEEE Conference on Computer Vision and Pattern
  Recognition, pp. 6790--6800 (2018)

\bibitem{mahdisoltani2018fine}
Mahdisoltani, F., Berger, G., Gharbieh, W., Fleet, D., Memisevic, R.:
  Fine-grained video classification and captioning.
\newblock arXiv preprint arXiv:1804.09235 \textbf{5}(6) (2018)

\bibitem{materzynska2020something}
Materzynska, J., Xiao, T., Herzig, R., Xu, H., Wang, X., Darrell, T.:
  Something-else: Compositional action recognition with spatial-temporal
  interaction networks.
\newblock In: Proceedings of the IEEE/CVF Conference on Computer Vision and
  Pattern Recognition, pp. 1049--1059 (2020)

\bibitem{plizzari2021skeleton}
Plizzari, C., Cannici, M., Matteucci, M.: Skeleton-based action recognition via
  spatial and temporal transformer networks.
\newblock Computer Vision and Image Understanding \textbf{208}, 103,219 (2021)

\bibitem{ranjan2017optical}
Ranjan, A., Black, M.J.: Optical flow estimation using a spatial pyramid
  network.
\newblock In: Proceedings of the IEEE conference on computer vision and pattern
  recognition, pp. 4161--4170 (2017)

\bibitem{ren2016faster}
Ren, S., He, K., Girshick, R., Sun, J.: Faster r-cnn: Towards real-time object
  detection with region proposal networks.
\newblock IEEE transactions on pattern analysis and machine intelligence
  \textbf{39}(6), 1137--1149 (2016)

\bibitem{simonyan2014two}
Simonyan, K., Zisserman, A.: Two-stream convolutional networks for action
  recognition in videos.
\newblock In: Advances in neural information processing systems, pp. 568--576
  (2014)

\bibitem{srinivas2021bottleneck}
Srinivas, A., Lin, T.Y., Parmar, N., Shlens, J., Abbeel, P., Vaswani, A.:
  Bottleneck transformers for visual recognition.
\newblock arXiv preprint arXiv:2101.11605  (2021)

\bibitem{sun2018pwc}
Sun, D., Yang, X., Liu, M.Y., Kautz, J.: Pwc-net: Cnns for optical flow using
  pyramid, warping, and cost volume.
\newblock In: Proceedings of the IEEE conference on computer vision and pattern
  recognition, pp. 8934--8943 (2018)

\bibitem{szegedy2017inception}
Szegedy, C., Ioffe, S., Vanhoucke, V., Alemi, A.A.: Inception-v4,
  inception-resnet and the impact of residual connections on learning.
\newblock In: Thirty-first AAAI conference on artificial intelligence (2017)

\bibitem{szegedy2015going}
Szegedy, C., Liu, W., Jia, Y., Sermanet, P., Reed, S., Anguelov, D., Erhan, D.,
  Vanhoucke, V., Rabinovich, A.: Going deeper with convolutions.
\newblock In: Proceedings of the IEEE conference on computer vision and pattern
  recognition, pp. 1--9 (2015)

\bibitem{szegedy2016rethinking}
Szegedy, C., Vanhoucke, V., Ioffe, S., Shlens, J., Wojna, Z.: Rethinking the
  inception architecture for computer vision.
\newblock In: Proceedings of the IEEE conference on computer vision and pattern
  recognition, pp. 2818--2826 (2016)

\bibitem{tian2020self}
Tian, Y., Che, Z., Bao, W., Zhai, G., Gao, Z.: Self-supervised motion
  representation via scattering local motion cues.
\newblock In: European Conference on Computer Vision, pp. 71--89. Springer
  (2020)

\bibitem{tian2021self}
Tian, Y., Lu, G., Min, X., Che, Z., Zhai, G., Guo, G., Gao, Z.:
  Self-conditioned probabilistic learning of video rescaling.
\newblock In: Proceedings of the IEEE/CVF International Conference on Computer
  Vision, pp. 4490--4499 (2021)

\bibitem{tian2019video}
Tian, Y., Min, X., Zhai, G., Gao, Z.: Video-based early asd detection via
  temporal pyramid networks.
\newblock In: 2019 IEEE International Conference on Multimedia and Expo, pp.
  272--277. IEEE (2019)

\bibitem{touvron2020training}
Touvron, H., Cord, M., Douze, M., Massa, F., Sablayrolles, A., J{\'e}gou, H.:
  Training data-efficient image transformers \& distillation through attention.
\newblock arXiv preprint arXiv:2012.12877  (2020)

\bibitem{tran2015learning}
Tran, D., Bourdev, L., Fergus, R., Torresani, L., Paluri, M.: Learning
  spatiotemporal features with 3d convolutional networks.
\newblock In: Proceedings of the IEEE international conference on computer
  vision, pp. 4489--4497 (2015)

\bibitem{tran2018closer}
Tran, D., Wang, H., Torresani, L., Ray, J., LeCun, Y., Paluri, M.: A closer
  look at spatiotemporal convolutions for action recognition.
\newblock In: Proceedings of the IEEE conference on Computer Vision and Pattern
  Recognition, pp. 6450--6459 (2018)

\bibitem{vaswani2017attention}
Vaswani, A., Shazeer, N., Parmar, N., Uszkoreit, J., Jones, L., Gomez, A.N.,
  Kaiser, L., Polosukhin, I.: Attention is all you need.
\newblock arXiv preprint arXiv:1706.03762  (2017)

\bibitem{wang2020video}
Wang, H., Tran, D., Torresani, L., Feiszli, M.: Video modeling with correlation
  networks.
\newblock In: Proceedings of the IEEE/CVF Conference on Computer Vision and
  Pattern Recognition, pp. 352--361 (2020)

\bibitem{wang2018appearance}
Wang, L., Li, W., Li, W., Van~Gool, L.: Appearance-and-relation networks for
  video classification.
\newblock In: Proceedings of the IEEE conference on computer vision and pattern
  recognition, pp. 1430--1439 (2018)

\bibitem{wang2020tdn}
Wang, L., Tong, Z., Ji, B., Wu, G.: Tdn: Temporal difference networks for
  efficient action recognition.
\newblock arXiv preprint arXiv:2012.10071  (2020)

\bibitem{wang2016temporal}
Wang, L., Xiong, Y., Wang, Z., Qiao, Y., Lin, D., Tang, X., Van~Gool, L.:
  Temporal segment networks: Towards good practices for deep action
  recognition.
\newblock In: European conference on computer vision, pp. 20--36. Springer
  (2016)

\bibitem{wang2018temporal}
Wang, L., Xiong, Y., Wang, Z., Qiao, Y., Lin, D., Tang, X., Van~Gool, L.:
  Temporal segment networks for action recognition in videos.
\newblock IEEE transactions on pattern analysis and machine intelligence
  \textbf{41}(11), 2740--2755 (2018)

\bibitem{wang2018non}
Wang, X., Girshick, R., Gupta, A., He, K.: Non-local neural networks.
\newblock In: Proceedings of the IEEE conference on computer vision and pattern
  recognition, pp. 7794--7803 (2018)

\bibitem{wang2018videos}
Wang, X., Gupta, A.: Videos as space-time region graphs.
\newblock In: Proceedings of the European conference on computer vision, pp.
  399--417 (2018)

\bibitem{wu2021coarse}
Wu, Z., Li, H., Zheng, Y., Xiong, C., Jiang, Y.G., Davis, L.S.: A
  coarse-to-fine framework for resource efficient video recognition.
\newblock International Journal of Computer Vision  (2021)

\bibitem{xie2018rethinking}
Xie, S., Sun, C., Huang, J., Tu, Z., Murphy, K.: Rethinking spatiotemporal
  feature learning: Speed-accuracy trade-offs in video classification.
\newblock In: Proceedings of the European Conference on Computer Vision, pp.
  305--321 (2018)

\bibitem{yang2019condconv}
Yang, B., Bender, G., Le, Q.V., Ngiam, J.: Condconv: Conditionally
  parameterized convolutions for efficient inference.
\newblock arXiv preprint arXiv:1904.04971  (2019)

\bibitem{zach2007duality}
Zach, C., Pock, T., Bischof, H.: A duality based approach for realtime tv-l 1
  optical flow.
\newblock In: Joint pattern recognition symposium, pp. 214--223. Springer
  (2007)

\bibitem{zhang2020pan}
Zhang, C., Zou, Y., Chen, G., Gan, L.: Pan: Towards fast action recognition via
  learning persistence of appearance.
\newblock arXiv preprint arXiv:2008.03462  (2020)

\bibitem{zhang2018shufflenet}
Zhang, X., Zhou, X., Lin, M., Sun, J.: Shufflenet: An extremely efficient
  convolutional neural network for mobile devices.
\newblock In: Proceedings of the IEEE conference on computer vision and pattern
  recognition, pp. 6848--6856 (2018)

\bibitem{zhang2021vidtr}
Zhang, Y., Li, X., Liu, C., Shuai, B., Zhu, Y., Brattoli, B., Chen, H., Marsic,
  I., Tighe, J.: Vidtr: Video transformer without convolutions.
\newblock In: Proceedings of the IEEE/CVF International Conference on Computer
  Vision, pp. 13,577--13,587 (2021)

\bibitem{zhou2018temporal}
Zhou, B., Andonian, A., Oliva, A., Torralba, A.: Temporal relational reasoning
  in videos.
\newblock In: Proceedings of the European Conference on Computer Vision, pp.
  803--818 (2018)

\bibitem{zolfaghari2018eco}
Zolfaghari, M., Singh, K., Brox, T.: Eco: Efficient convolutional network for
  online video understanding.
\newblock In: Proceedings of the European conference on computer vision, pp.
  695--712 (2018)

\end{thebibliography}
}

\end{document}